\documentclass[10pt,twocolumn,letterpaper]{article}

\usepackage[pagenumbers]{cvpr} 

\usepackage[table,xcdraw,dvipsnames]{xcolor}
\makeatletter
\@namedef{ver@everyshi.sty}{}
\newcommand\newtag[2]{#1\def\@currentlabel{#1}\label{#2}}
\makeatother
\usepackage{svg}
\usepackage{graphicx}
\usepackage{amsmath}
\usepackage{amssymb}
\usepackage{booktabs}
\usepackage{xspace}
\usepackage{multirow}
\usepackage[normalem]{ulem}
\useunder{\uline}{\ul}{}
\usepackage{bm}
\usepackage[accsupp]{axessibility}

\newcommand{\method}{PDV\xspace}
\def\*#1{\mathbf{#1}}

\usepackage[pagebackref,breaklinks,colorlinks]{hyperref}

\usepackage[capitalize]{cleveref}
\crefname{section}{Sec.}{Secs.}
\Crefname{section}{Section}{Sections}
\Crefname{table}{Table}{Tables}
\crefname{table}{Tab.}{Tabs.}

\def\cvprPaperID{9743} 
\def\confName{CVPR}
\def\confYear{2022}

\begin{document}

\title{Point Density-Aware Voxels for LiDAR 3D Object Detection}

\author{
Jordan S. K. Hu \ \ \ \ Tianshu Kuai \ \ \ \ Steven L. Waslander\\
University of Toronto Robotics Institute\\
{\tt\small \{jordan.hu, tianshu.kuai\}@mail.utoronto.ca, steven.waslander@robotics.utias.utoronto.ca }
}
\maketitle

\begin{abstract}
LiDAR has become one of the primary 3D object detection sensors in autonomous driving. However, LiDAR's diverging point pattern with increasing distance results in a non-uniform sampled point cloud ill-suited to discretized volumetric feature extraction. Current methods either rely on voxelized point clouds or use inefficient farthest point sampling to mitigate detrimental effects caused by density variation but largely ignore point density as a feature and its predictable relationship with distance from the LiDAR sensor. Our proposed solution, Point Density-Aware Voxel network (PDV), is an end-to-end two stage LiDAR 3D object detection architecture that is designed to account for these point density variations. PDV efficiently localizes voxel features from the 3D sparse convolution backbone through voxel point centroids. The spatially localized voxel features are then aggregated through a density-aware RoI grid pooling module using kernel density estimation (KDE) and self-attention with point density positional encoding. Finally, we exploit LiDAR's point density to distance relationship to refine our final bounding box confidences. PDV outperforms all state-of-the-art methods on the Waymo Open Dataset and achieves competitive results on the KITTI dataset. We provide a code release for \method which is available at \url{https://github.com/TRAILab/PDV}.
\end{abstract}

\section{Introduction}
\label{sec:intro}

\begin{figure}[t!]
\centering
\begin{tabular}{ccc}
    \includegraphics[width=0.28\columnwidth]{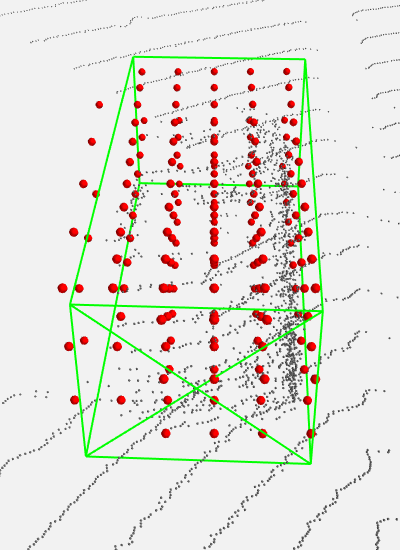} &
    \includegraphics[width=0.28\columnwidth]{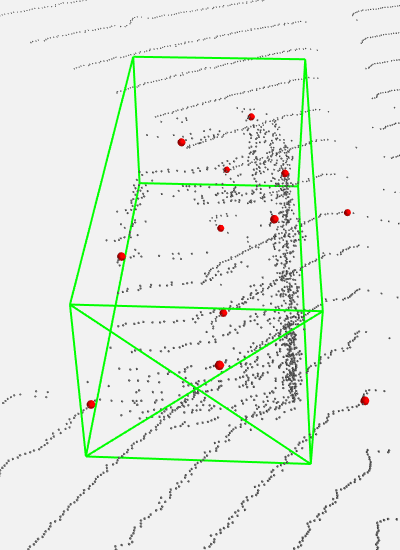} &
    \includegraphics[width=0.28\columnwidth]{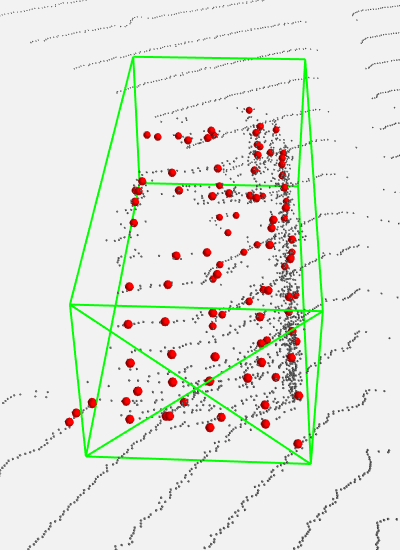}\\
    (a) & (b) & (c)
\end{tabular}
\caption{Voxel feature localization using (a)~voxel centers, (b)~farthest point sampling, and (c)~voxel point centroids on a vehicle in the Waymo Open Dataset \cite{sun2020scalability}. By using the raw point cloud to localize voxel features, voxel point centroids provide dense geometric information for second-stage proposal refinement.}
\label{fig:intro}
\end{figure}

3D object detection is one of the key perception problems in the autonomous vehicle space as object pose estimation directly impacts the effectiveness of downstream tasks in the perception pipeline. Within the autonomous driving sensor stack, LiDAR has become one of the most popular sensors used for 3D object detection~\cite{shi2020pv, second, Shi_2019_CVPR_pointrcnn}, because of the accurate 3D point cloud it produces through laser light.

However, the reliance on LiDAR data comes at the cost of point density variations across distance. Other factors such as occlusion play a role, but the primary reason is the natural divergence of points from the LiDAR with increasing distance due to the angular offsets between the LiDAR lasers. Thus, objects located at farther distances return fewer points than objects located closer to the LiDAR.

Voxel-based methods~\cite{voxelnet, second, zheng2020ciassd, deng2020voxelrcnn} typically ignore point density, solely relying on the quantized representation of the point cloud. When a high voxel resolution is afforded, as is the case on the KITTI dataset~\cite{geiger2012vision}, voxel-based methods~\cite{sessd} have outperformed point-based and point-voxel-based methods. However, on datasets with larger input spaces such as the Waymo Open Dataset~\cite{sun2020scalability}, the voxel resolution is limited due to memory constraints. Fine object details are therefore lost due to spatial misalignment between the voxel features and the point cloud as shown in \Cref{fig:intro}~(a), resulting in a degradation in performance.

Other methods~\cite{Shi_2019_CVPR_pointrcnn, shi2020pv} attempt to remedy point density variations through farthest point sampling (FPS) as seen in \Cref{fig:intro}~(b). Although effective at sampling locations on non-uniformly distributed point clouds, the computation scales poorly as a function of the number of points in the point cloud, increasing runtime and limiting the number of sampled points for second-stage proposal refinement.

Point density also affects detection of smaller objects such as pedestrians and cyclists. These objects have less surface area to intersect the LiDAR's laser beams, resulting in poorer object localization. Perhaps informatively, current state-of-the-art methods have largely ignored detection performance for pedestrians and cyclists, focusing solely on the car or vehicle class~\cite{deng2020voxelrcnn, Mao_2021_ICCV_pyramidrcnn, Mao_2021_ICCV_votr, sessd}. As we move towards datasets with higher environment coverage, it is necessary for architectures to be scalable to larger input spaces and to serve as a multi-class solution for 3D object detection.

We therefore propose Point Density-Aware Voxel network (\method) to resolve these identified issues by leveraging voxel point centroid localization and feature encodings that directly account for point density in multi-class 3D object detection. We summarize our approach with the following three contributions.

\noindent\textbf{(1) Voxel Point Centroid Localization.} \method partitions the LiDAR points in each non-empty voxel to calculate a point centroid for each voxel feature, as shown in \Cref{fig:intro}~(c). By localizing the voxel features with point centroids for second-stage proposal refinement, \method uses the point density distributions to retain fine-grained position information in the feature encodings without requiring an expensive point cloud sampling method such as FPS.

\noindent\textbf{(2) Density-Aware RoI Grid Pooling.} We augment region of interest (RoI) grid pooling~\cite{shi2020pv} to encode local point density as an additional feature. First, we use kernel density estimation (KDE)~\cite{parzen1962pdf, rosenblatt1956nonpar} to encode local voxel feature density at each grid point ball query, followed by self-attention~\cite{vaswani2017attention} between grid points with a novel point density positional encoding. Density-aware RoI grid pooling captures localized point density information in the context of the whole region proposal for second-stage refinement.

\noindent\textbf{(3) Density Confidence Prediction.} We further refine our bounding box confidence predictions by using the final bounding box centroid location and the number of raw LiDAR points within the final bounding box as additional features. Thus, we use the inherent relationship between distance and point density established by LiDAR for more informed confidence predictions.

\method outperforms all current state-of-the-art methods on the Waymo Open Dataset~\cite{sun2020scalability} with an increase of +0.65\%/+1.25\%, +0.53\%/+0.46\%, and +0.49\%/+0.71\% on the vehicle, pedestrian, and cyclist LEVEL\_1/LEVEL\_2 mAPH classes, respectively, and achieves competitive performance on the KITTI dataset~\cite{geiger2012vision}.

\section{Related Work}
\label{sec:rel-work}

\begin{figure*}[t!]
  \centering
  \includegraphics[width=\textwidth]{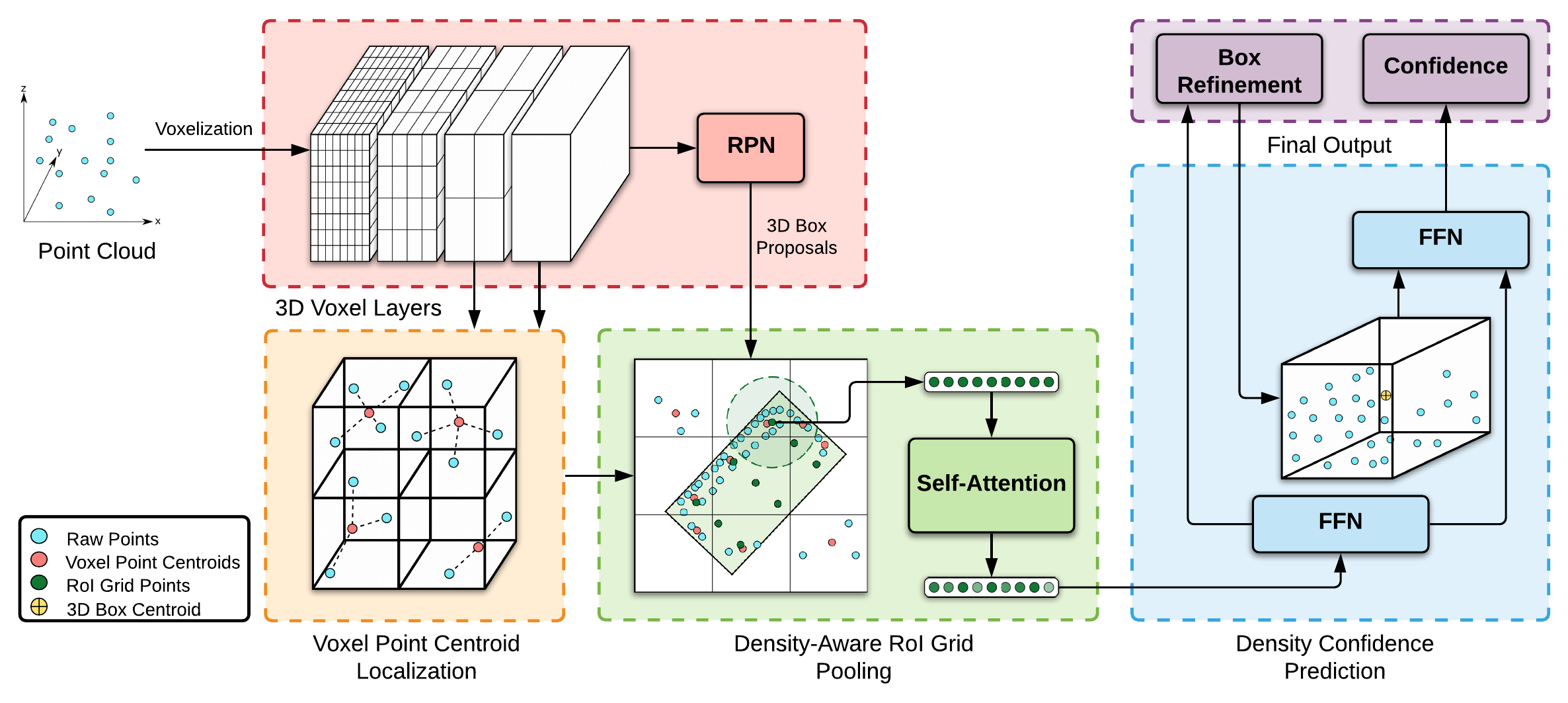}

  \caption{\method architecture. The input point cloud is first voxelized and processed through 3D sparse convolutions and a RPN head to produce initial bounding box proposals. Voxel features in each layer $l=1,\dots, L$ are then localized via voxel point centroids, which are then aggregated through density-aware RoI grid pooling. The RoI grid features are used to refine the bounding box proposals and their associated confidence, with additional adjustments from density confidence prediction.}
  \label{fig:full-diagram}
\end{figure*}

\noindent\textbf{Point-based LiDAR 3D Object Detection.} Point-based methods use the raw point cloud to extract point-level features for bounding box predictions. F-PointNet~\cite{Qi_2018_CVPR} applies PointNet~\cite{qi2017pointnet, qi2017pointnet++} on the point cloud, segmented via image-based 2D object detections. PointRCNN~\cite{Shi_2019_CVPR_pointrcnn} directly generates RoIs at the point-level through a PointNet++\cite{qi2017pointnet++} backbone and uses point-level features for bounding box refinement. STD~\cite{yang2019std} proposes PointsPool for RoI feature extraction while 3DSSD~\cite{Yang20203dssd} adopts a new sampling strategy on the raw point cloud to preserve enough interior points for objects in the downsampled point cloud. Point-GNN\cite{shi2020point} constructs a graph using the raw point cloud and aggregates node-level features to generate predictions. Point-based methods utilize expensive point cloud sampling and grouping, which inevitably require long inference times.

\noindent\textbf{Voxel-based LiDAR 3D Object Detection.} Voxel-based methods divide the point cloud into a voxel grid to directly apply 3D and 2D convolutions for generating predictions~\cite{voxelnet,second,lang2019pointpillars}. CIA-SSD~\cite{zheng2020ciassd} adopts a light network on the bird's eye biew (BEV) grid to extract robust spatial-semantic features with a confidence rectification module for better post-processing. Voxel-RCNN~\cite{deng2020voxelrcnn} proposes Voxel RoI pooling to generate RoI features by aggregating voxel features. VoTr~\cite{Mao_2021_ICCV_votr} proposes a transformer-based 3D backbone as an alternative to the standard sparse convolution layers~\cite{second, graham20183d}. The performance of voxel-based methods is limited by the quantized point cloud as fine-grained point-level information is lost from the voxelization process.

\noindent\textbf{Point-Voxel-based LiDAR 3D Object Detection.} Point-voxel-based methods utilize both voxel and point representations of the point cloud. SA-SSD~\cite{he2020sassd} uses an auxiliary network during training that interpolates point-level features from intermediate voxel layers. PV-RCNN~\cite{shi2020pv} adopts RoI grid pooling to effectively aggregate FPS-sampled keypoint features on evenly spaced grids inside each bounding box proposal. PV-RCNN++~\cite{shi2021pv} proposes a modified version of FPS for faster point sampling and VectorPool aggregation for RoI grid pooling. CT3D~\cite{Sheng_2021_ICCV_ct3d} constructs a cylindrical-shaped RoI at each bounding box proposal. It adopts a transformer-based encoder-decoder architecture to extract RoI features directly from the nearby points without using intermediate voxel features. Pyramid-RCNN~\cite{Mao_2021_ICCV_pyramidrcnn} extends the idea of RoI grid pooling over a single set of evenly spaced grids to multiple sets of grids at different scales with adaptive ball query radius at a significantly higher computation cost. Current point-voxel based methods do not explicitly account for variations in point cloud density within each RoI and typically require long inference times due to the dependence on point cloud sampling.

\noindent\textbf{Point Density Estimation.} KDE~\cite{parzen1962pdf, rosenblatt1956nonpar} estimates a probability distribution function of a random variable using a finite set of samples and a chosen kernel function and bandwidth. A couple methods have used KDE for feature encoding in point clouds. MC Convolution~\cite{hermosilla2018monte} uses a Monte Carlo estimation of the convolution integral to handle non-uniform sampled point clouds and uses KDE to estimate the likelihood of a point within a local convolution. PointConv~\cite{wu2019pointconv} also uses KDE, but estimates the likelihood of each sample with an additional feed-forward network (FFN). Rather than restricting the density estimate for reweighting, we use KDE as an additional feature within each grid point ball query in density-aware RoI grid pooling.

\section{Methodology}

\label{sec:method}
\method uses a two-stage approach with a 3D sparse convolution backbone for initial bounding box proposals which are then refined in a second stage through voxel features in each voxel layer and the raw point cloud data. \Cref{fig:full-diagram} shows an overview of the \method framework.

\subsection{3D Voxel Backbone}
We use a similar voxel backbone for initial bounding box proposals to SECOND~\cite{second}. The input to \method is a point cloud which is defined as a set of 3D points $\left\{\*p_i = \left\{\*x_{\*p_i}, \*f_{\*p_i}\right\} \, | \, i=1,\dots,N_\*p\right\}$ where $\*x_{\*p_i} \in \mathbb{R}^{3}$ are the xyz spatial coordinates, $\*f_{\*p_i} \in \mathbb{R}^{F}$ are additional features such as  the intensity or elongation of each point, and $N_\*p$ is the number of points in the point cloud. First, the point cloud is voxelized and subsequently encoded using a series of 3D sparse convolutions~\cite{second, graham20183d}, followed by a region proposal network (RPN) for initial bounding box proposals. Each voxel layer has a different spatial resolution with 1x, 2x, 4x, and 8x downsampled resolutions based on the original voxel grid size. The voxel features in each layer are used for bounding box refinement in the second stage.

\subsection{Voxel Point Centroid Localization}
Inspired by grid-subsampling in KPConv~\cite{thomas2018semantic, thomas2019kpconv}, the voxel point centroid localization module spatially locates non-empty voxel features for aggregation in density-aware RoI grid pooling.

\begin{figure}[t!]
  \centering
  \includegraphics[width=\linewidth]{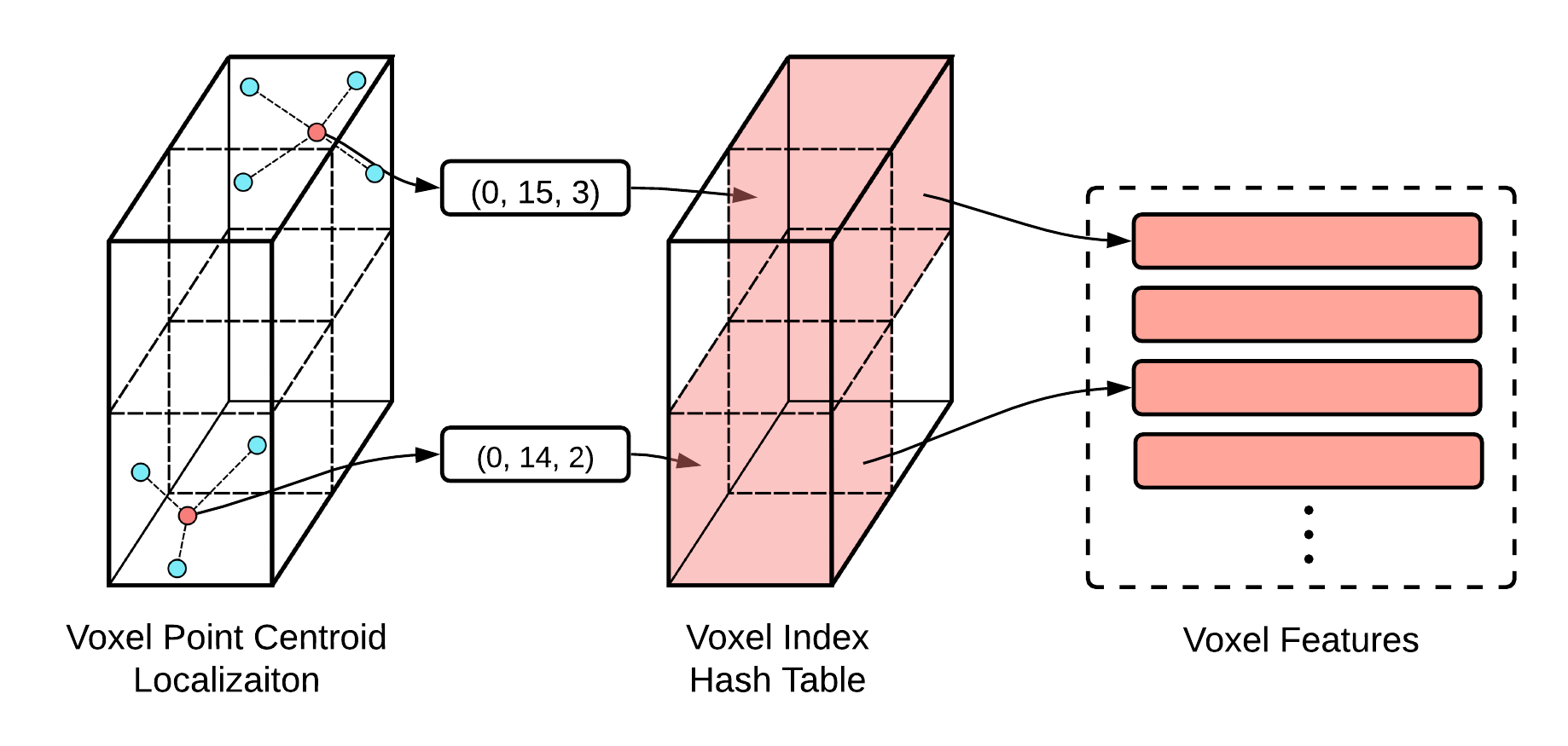}
  \caption{Voxel point centroids (\textcolor{RedOrange}{red}) are assigned to their respective voxel grid index. A 3D hash table maps the voxel index to the associated voxel feature in the sparse convolution layer.}
  \label{fig:voxel-centroids}
\end{figure}

Let $\bm{\mathcal{V}}^l = \{\*V^l_k = \{\*h_{\*V^l_k}, \*f_{\*V^l_k}\} \, | \, k=1, \dots, N_l\}$ be the set of non-empty voxels in the $l$-th voxel layer where $\*h_{\*V^l_k}$ is the 3D voxel index, $\*f_{\*V^l_k}$ is the associated voxel feature vector, and $N_l$ is the number of non-empty voxels for voxel layers $l=1, \dots, L$. First, points that are within the same voxel are grouped together into a set $\mathcal{N}(\*V^l_k)$ by calculating their voxel index $\*h_{\*V^l_k}$ from their spatial coordinates $\*x_i$ and voxel grid dimensions. The point centroid of each voxel feature is then calculated as
\begin{equation}
    \*c_{\*V^l_k} = \frac{1}{|\mathcal{N}(\*V^l_k)|} \sum_{\*x_{\*p_i} \in \mathcal{N}(\*V^l_k)} \*x_{\*p_i}.
    \label{eq:important}
\end{equation}
Since the voxels in the convolution layers are in a sparse format, an intermediate hash table is used to efficiently map each calculated voxel point centroid to its corresponding feature vector. As shown in \Cref{fig:voxel-centroids}, both voxel point centroids and sparse voxel features are associated with a shared voxel index. The intermediate hash table links the centroid $\*c_{\*V^l_k}$ with $\*V^l_k$ using the matching voxel index $\*h_{\*V^l_k}$.

An advantage with using voxels is that we can use the previous voxel layer centroid calculations to efficiently compute the subsequent voxel point centroids based on the stride, padding, and kernel size of the convolution block. Let $\mathcal{C}^{l+1}_k = \{ \*c_{\*V^l_j} \, | \, \mathcal{K}_{l+1}(\*h_{\*V^l_j}) = \*h_{\*V^{l+1}_k} \}$ be the set of voxel point centroids where $\mathcal{K}_{l+1}$ is the convolution block that maps voxel index $\*h_{\*V^l_j}$ to $\*h_{\*V^{l+1}_k}$. We can then perform a weighted average of the grouped voxel point centroids to calculate the centroids in the subsequent layer:
\begin{equation}
    \*c_{\*V^{l+1}_k} = \frac{1}{\big|\mathcal{N}(\*V^{l+1}_k)\big|} \sum_{\*c_{\*V^l_j} \in \, \mathcal{C}^{l+1}_k} \big|\mathcal{N}(\*V^l_j)\big| \, \*c_{\*V^l_j}
\end{equation}
where $\big|\mathcal{N}(\*V^{l+1}_k)\big|$ is calculated from
\begin{equation}
    \big|\mathcal{N}(\*V^{l+1}_k)\big| = \sum_{\*c_{\*V^l_j} \in \, \mathcal{C}^{l+1}_k} \big|\mathcal{N}(\*V^l_j)\big|.
\end{equation}
By avoiding recomputing centroids using the entire point cloud for each layer, voxel point centroid localization scales to larger point clouds more efficiently.

\subsection{Density-aware RoI Grid Pooling}

Density-aware RoI grid pooling builds upon RoI grid pooling~\cite{shi2020pv} by augmenting the pooling method with a combination of KDE and self-attention to encode point density features into each proposal. First, $U \times U \times U$ uniform grid points $\bm{\mathcal{G^\*b}} = \{\*g_1, \dots, \*g_{U^3}\}$ are sampled for each bounding box proposal $\*b$.

\begin{figure}[t!]
\centering
\begin{tabular}{cc}
    \includegraphics[width=0.45\columnwidth]{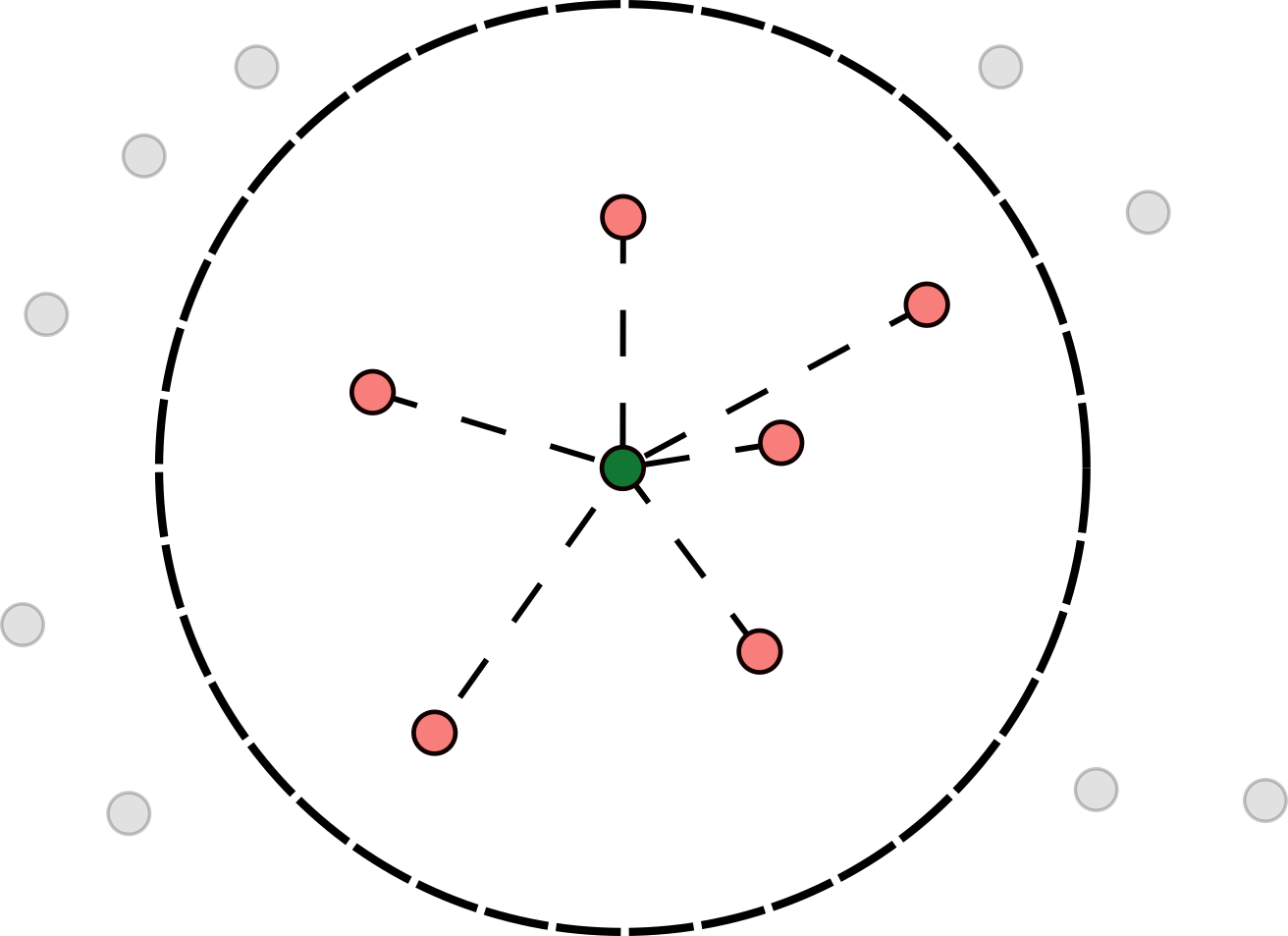} &
    \includegraphics[width=0.45\columnwidth]{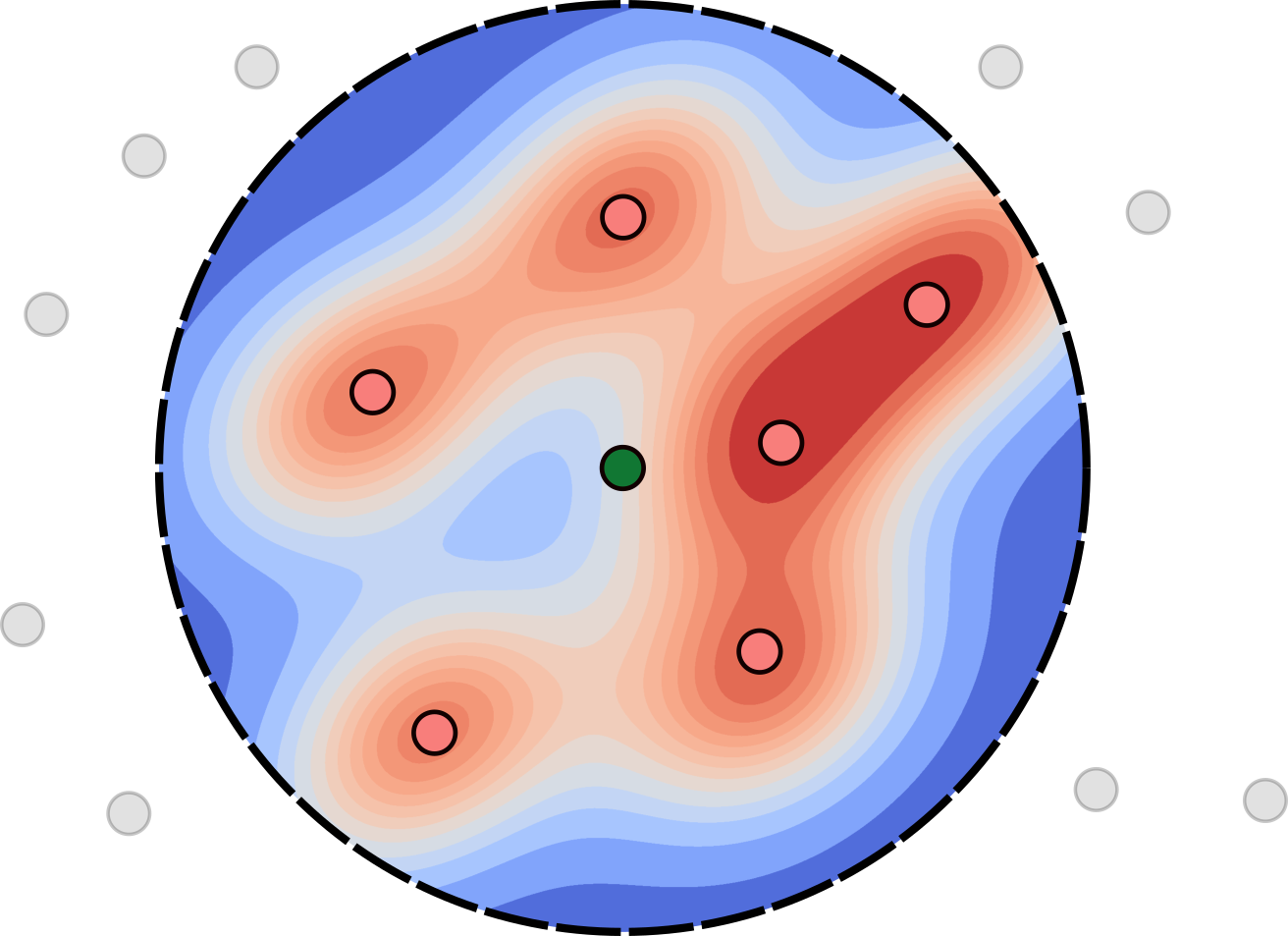}\\
    (a) & (b)
\end{tabular}
\caption{Two types of features are added to each grid point ball query: (a) the relative offset from the ball query center, and (b) the probability density of each point calculated through KDE. \textcolor{red}{Red} and \textcolor{blue}{blue} indicate higher and lower probability density, respectively.}
\label{fig:kde-ball-query}
\end{figure}

\noindent\textbf{Local Feature Density.} We use KDE to estimate local feature density within each grid point ball query. Rather than limiting the estimated density as a feature reweighting like MC Convolution~\cite{hermosilla2018monte} and PointConv~\cite{wu2019pointconv}, density-aware RoI grid pooling encodes the estimated probability density as an additional feature in the ball query for a more implicit feature encoding. First, we aggregate neighbouring features near each grid point where $\mathcal{N}(\*g_j)$ is the set of voxel point centroids in a sphere of radius $r$ centered around $\*g_j$:
\begin{equation}
    \Psi^l_{\*g_j} =
    \left\{
    \begin{bmatrix}
    \*f_{\*V^l_k}\\
    \*c_{\*V^l_k} - \*g_j\\
    p(\*c_{\*V^l_k} | \*g_j)
    \end{bmatrix}^\top, \, \forall \*c_{\*V^l_k} \in \mathcal{N}(\*g_j)
    \right\}
\end{equation}
where the local offset $\*c_{\*V^l_k} - \*g_j$ and likelihood $p(\*c_{\*V^l_k} | \*g_j)$ are appended as additional features, as shown in \Cref{fig:kde-ball-query}. The likelihood is calculated for each grid point using KDE:
\begin{equation}
    p(\*c_{\*V^l_k} | \*g_j) \approx \frac{1}{|\mathcal{N}(\*g_j)|\sigma^3} \sum_{\*c_{\*V^l_i} \in \mathcal{N}(\*g_j)} \mathcal{W}(\*c_{\*V^l_k}, \*c_{\*V^l_i})
\label{eq:kde-likelihood}
\end{equation}
where $\sigma$ is the bandwidth and $\mathcal{W}$ is
\begin{equation}
    \mathcal{W}(\*c_{\*V^l_k}, \*c_{\*V^l_i}) = \prod_{d=1}^3 w \Big(\frac{\*c_{\*V^l_k, d} - \*c_{\*V^l_i, d}}{\sigma}\Big)
\end{equation}
with an independent kernel $w$ on each xyz dimension $d$. Once the features are appended, a PointNet multi-scale grouping (MSG) module~\cite{qi2017pointnet, qi2017pointnet++} is used to obtain a feature vector $\*f^l_{\*g_j}$ for each grid point $\*g_j$:
\begin{equation}
    \*f^l_{\*g_j} = \textrm{maxpool} (\textrm{FFN}(\Psi^l_{\*g_j})).
\end{equation}
 We use multiple radii $r$ to capture feature density at different scales for each grid point and concatenate the output features together. Finally, features are appended from different voxel layers to obtain the final features for each grid point:
\begin{equation}
    \*f_{\*g_j} = \big[{\*f^{1}_{\*g_j}}, \dots, \*f^{L}_{\*g_j} \big].
\end{equation}

\noindent\textbf{Grid Point Self-Attention.} The features encoded at each RoI grid point are localized to the size of the ball query but lack interdependent relationships between different grid points. An easy solution is to use self-attention~\cite{vaswani2017attention} to capture long-range dependencies between the grid points but simply adding an attention module lacks the geometric information of the LiDAR point cloud. Thus, we also introduce a novel type of positional encoding that takes into consideration the point density within the point cloud.

\begin{figure}[!t]
  \centering
  \includegraphics[width=\linewidth]{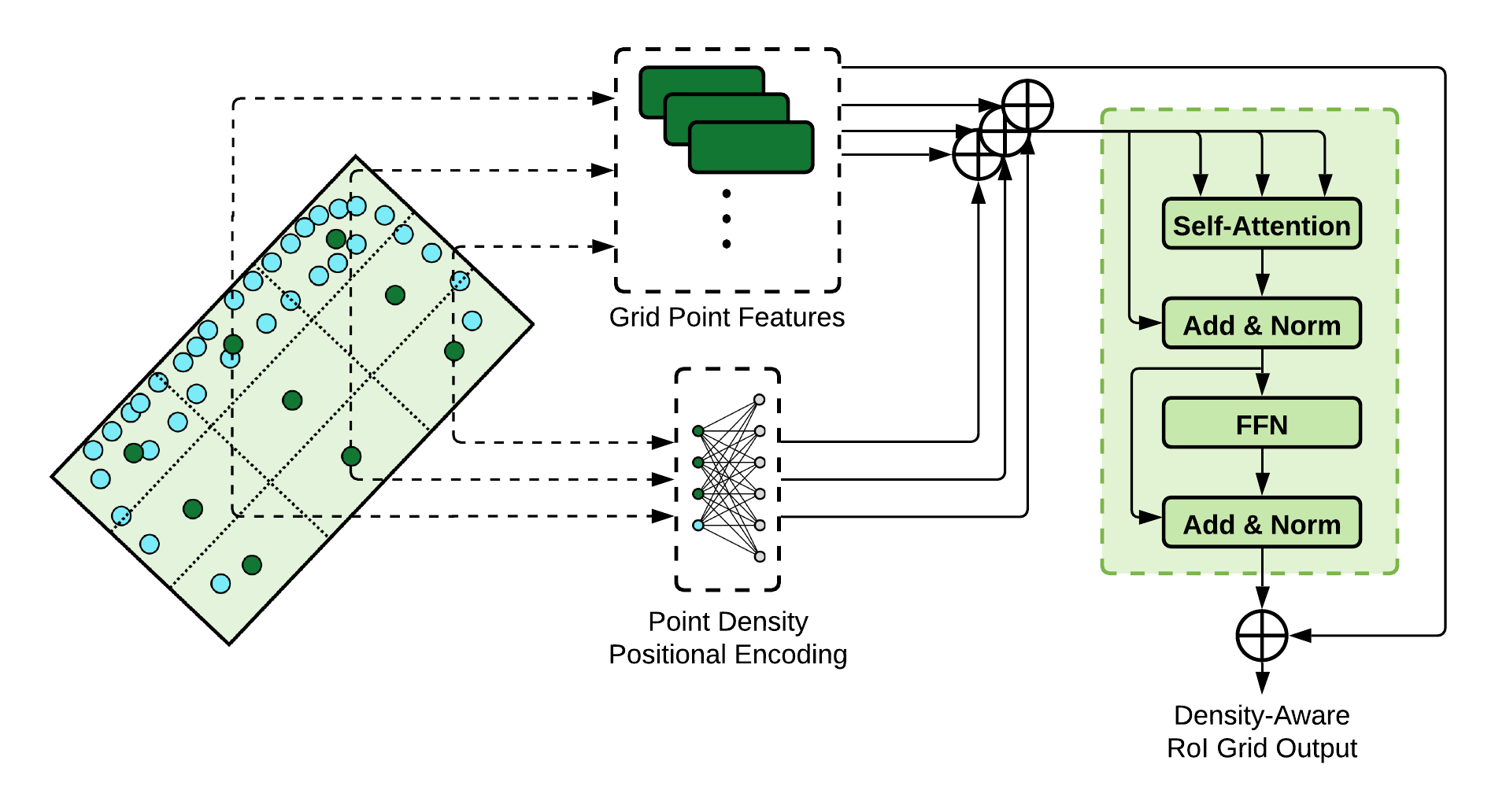}
  \caption{Density-aware RoI grid pooling performs self-attention on the grid point features (\textcolor{ForestGreen}{green}) to capture long-range dependencies. Point density positional encoding uses the relative offset  of the grid point and number of points (\textcolor{CornflowerBlue}{blue}) in each grid voxel as inputs.}
  \label{fig:attention}
\end{figure}

As shown in \Cref{fig:attention}, the self-attention module performs self-attention between the non-empty grid point features $\*f_{\mathcal{G}^\*b} = \left\{ \*f_{\*g_i} \, \Big| \, |\mathcal{N}(\*g_i)| > 0, \, \forall \*g_i \in \mathcal{G^\*b} \right\}$ using a standard transformer encoder layer~\cite{vaswani2017attention} and a residual connection similar to a non-local neural network block~\cite{wang2018non}:
\begin{equation}
\tilde{\*f}_{\*g_i} = \mathcal{T}_{\*g_i}(\*f_{\mathcal{G}^\*b}) + \*f_{\*g_i}
\end{equation}
where $\mathcal{T}_{\*g_i}$ is the transformer encoder layer output for $\*f_{\*g_i}$ and $\tilde{\*f}_{\*g_i}$ is the output grid feature. Empty grid point features $|\mathcal{N}(\*g_i)| = 0$ are untouched by the self-attention module and are kept as their original feature encoding.

\noindent\textbf{Point Density Positional Encoding.} We add positional encoding to the self-attention module by using the local grid point positions and the number of points in the box proposal. The bounding box proposal is divided into voxels $\*V_{\*g_j}$ using the same $U\times U \times U$ grid resolution to establish voxels for each grid point. The positional encoding for each grid feature is then calculated as:
\begin{equation}
    \textrm{PE}(\*f_{\*g_j}) = \textrm{FFN}\left(\left[\bm{\delta}_{\*g_j}, \, \textrm{log}\left(\left|\mathcal{N}\left(\*V_{\*g_j}\right)\right| + \epsilon\right)\right]\right)
\end{equation}
where $\bm{\delta}_{\*g_j} =\*x_{\*g_j} - \*c_\*b$ is the relative position of $\*g_j$ from the bounding box proposal centroid $\*c_\*b$,  $|\mathcal{N}(\*V_{\*g_j})|$ is the number of points in each grid point voxel $\*V_{\*g_j}$, and $\epsilon$ is a constant offset. By leveraging the local offsets and number of points within each voxel, density-aware RoI grid pooling is able to capture point densities within each region proposal.

\subsection{Density Confidence Prediction}
\method also uses the relationship between the distance and the number of LiDAR points on scanned objects to predict the confidence of the final bounding box predictions. A shared FFN first encodes the flattened features from the density-aware RoI grid pooling module. Then, two separate FFN branches encode the features for the box refinement and box confidence outputs. In the box confidence branch, we additionally append two features to predict the output confidence $p_{\tilde{\*b}}$ for the final bounding box $\tilde{\*b}$:
\begin{equation}
    p_{\tilde{\*b}} = \textrm{FFN}\Big(\Big[\*f^s_{\tilde{\*b}}, \, \*c_{\tilde{\*b}}, \, \textrm{log}\big(|\mathcal{N}(\tilde{\*b})|\big)\Big]\Big)
\end{equation}
where $\*f^s_{\tilde{\*b}}$ is the output feature vector from the shared FFN, $\*c_{\tilde{\*b}}$ is the centroid of the final bounding box, and $|\mathcal{N}(\tilde{\*b})|$ is the number of raw points in the final bounding box.

\subsection{Training Losses}
We use an end-to-end training strategy for \method with a region proposal loss $L_{\textrm{RPN}}$ and proposal refinement loss $L_{\textrm{RCNN}}$ that are trained jointly. The $L_{\textrm{RPN}}$ is calculated as
\begin{equation}
    L_{\textrm{RPN}} = L_{\textrm{cls}}(\*y_\*b, \*y^\star_\*b) + \beta L_{\textrm{reg}}(\*r_\*b, \*r^\star_\*b)
\end{equation}
where $L_{\textrm{cls}}$ is the focal loss \cite{lin2017focal}, $L_{\textrm{reg}}$ is the smooth-L1 loss, $\*y_\*b$ is the predicted class vector, $\*y^\star_\*b$ is the ground truth class, $\*r_\*b$ is the predicted RoI anchor residual, $\*r^\star_\*b$ is the ground truth anchor residual, and $\beta$ is a scaling factor. $L_{\textrm{RCNN}}$ is composed of
\begin{equation}
    L_{\textrm{IoU}} = -p^\star_{\tilde{\*b}} \textrm{log}(p_{\tilde{\*b}}) - (1 - p^\star_{\tilde{\*b}}) \textrm{log}(1 - p_{\tilde{\*b}})
\end{equation}
where $p^\star_{\tilde{\*b}}$ is the confidence training target scaled by the 3D RoI and their associated ground truth bounding box as done in PV-RCNN \cite{shi2020pv}. Thus $L_{\textrm{RCNN}}$ is
\begin{equation}
    L_{\textrm{RCNN}} = L_{\textrm{IoU}} + L_{\textrm{reg}}(\*r_{\tilde{\*b}}, \*r^\star_{\tilde{\*b}})
\end{equation}
where $\*r_{\tilde{\*b}}$ is the predicted bounding box residual and $\*r^\star_{\tilde{\*b}}$ is the ground truth residual. A smooth-L1 loss is used to regress the bounding box residuals. We use the same confidence and regression targets as PV-RCNN~\cite{shi2020pv}.
\section{Experimental Results}
\label{ssec:exp-setup}

\begin{table*}[ht!]
\centering
\small
\resizebox{\textwidth}{!}{
\begin{tabular}{l|cccc|cccc|cccc}
\hline
                         & \multicolumn{2}{c}{Veh. (LEVEL\_1)} & \multicolumn{2}{c|}{Veh. (LEVEL\_2)} & \multicolumn{2}{c}{Ped. (LEVEL\_1)} & \multicolumn{2}{c|}{Ped. (LEVEL\_2)} & \multicolumn{2}{c}{Cyc. (LEVEL\_1)} & \multicolumn{2}{c}{Cyc. (LEVEL\_2)} \\
\multirow{-2}{*}{Method} & mAP              & mAPH             & mAP               & mAPH             & mAP              & mAPH             & mAP               & mAPH             & mAP              & mAPH             & mAP              & mAPH             \\ \hline
SECOND \cite{second}                   & 72.27            & 71.69            & 63.85             & 63.33            & 68.7             & 58.18            & 60.72             & 51.31            & 60.62            & 59.28            & 58.34            & 57.05            \\
PointPillar \cite{lang2019pointpillars}              & 56.62            & -                & -                 & -                & 59.25                & -                & -                 & -                & -                & -                & -                & -                \\
$\textrm{MVF}^\star$ \cite{zhou2020end}                     & 62.93            & -                & -                 & -                & 65.33            & -                & -                 & -                & -                & -                & -                & -                \\
$\textrm{Pillar-OD}^\star$ \cite{wang2020pillar}               & 69.8             & -                & -                 & -                & 72.51            & -                & -                 & -                & -                & -                & -                & -                \\
AFDet \cite{ge2020afdet}                    & 63.69            & -                & -                 & -                & -                & -                & -                 & -                & -                & -                & -                & -                \\
$\textrm{Part-A2-Net}^\dagger$ \cite{shi2020parta2}              & 74.82            & 74.32            & 65.88             & 65.42            & 71.76            & 63.64            & 62.53             & 55.3             & 67.35            & 66.15            & 65.05            & 63.89            \\
$\textrm{PV-RCNN}^\dagger$ \cite{shi2020pv}                  & 75.17            & 74.6             & 66.35             & 65.84            & 72.65            & 63.52            & 63.42             & 55.29            & 67.26            & 65.82            & 64.88            & 63.48            \\
PV-RCNN++ \cite{shi2021pv}                & 76.14            & 75.62            & 68.05             & 67.56            & 73.97            & 65.43            & 65.64             & 57.82            & 68.38            & 67.06            & 65.92            & 64.65            \\
Voxel-RCNN \cite{deng2020voxelrcnn}               & 75.59            & -                & 66.59             & -                & -                & -                & -                 & -                & -                & -                & -                & -                \\
CT3D \cite{Sheng_2021_ICCV_ct3d}                     & 76.3             & -                & 69.04             & -                & -                & -                & -                 & -                & -                & -                & -                & -                \\
VoTr-TSD \cite{Mao_2021_ICCV_votr}                 & 74.95            & 74.25            & 65.91             & 65.29            & -                & -                & -                 & -                & -                & -                & -                & -                \\
Pyramid-PV \cite{Mao_2021_ICCV_pyramidrcnn}               & 76.3             & 75.68            & 67.23             & 66.68            & -                & -                & -                 & -                & -                & -                & -                & -                \\ \hline
PDV (Ours)               & \textbf{76.85}   & \textbf{76.33}   & \textbf{69.30}    & \textbf{68.81}   & \textbf{74.19}   & \textbf{65.96}   & \textbf{65.85}    & \textbf{58.28}   & \textbf{68.71}   & \textbf{67.55}   & \textbf{66.49}   & \textbf{65.36}   \\
\rowcolor[HTML]{E0FFFF} 
\textit{Improvement}     & \textit{+0.55}   & \textit{+0.65}   & \textit{+0.26}    & \textit{+1.25}   & \textit{+0.22}   & \textit{+0.53}   & \textit{+0.21}    & \textit{+0.46}   & \textit{+0.33}   & \textit{+0.49}   & \textit{+0.57}   & \textit{+0.71}   \\ \hline
\end{tabular}
}
\caption{Performance comparison on the Waymo Open Dataset with 202 validation sequences for 3D vehicle (IoU~=~0.7), pedestrian (IoU~=~0.5) and cyclist (IoU~=~0.5) detection. $\star$: Results are on Waymo Open Dataset 1.0 version. $\dagger$: Results are from~\cite{shi2021pv}.}
\label{tab:waymo-val}
\end{table*}

\begin{table}[!ht]
\centering
\small
\resizebox{\columnwidth}{!}{
\begin{tabular}{l|ccc}
\hline
\multirow{2}{*}{Method} & \multicolumn{3}{c}{LEVEL 1 mAP/mAPH}                           \\
                        & 0-30m                & 30-50m               & 50m-Inf          \\ \hline
$\textrm{PV-RCNN}^\star$ \cite{shi2020pv}                & 91.92/91.34          & 69.21/68.53          & 42.17/41.31      \\
Voxel-RCNN \cite{deng2020voxelrcnn}             & 92.49/-              & 74.09/-              & 53.15/-          \\
CT3D \cite{Sheng_2021_ICCV_ct3d}                    & 92.51/-              & 75.07/-              & \textbf{55.36}/- \\
VoTr-TSD \cite{Mao_2021_ICCV_votr}                & 92.28/91.73          & 73.36/72.56          & 51.09/50.01      \\
Pyramid-PV \cite{Mao_2021_ICCV_pyramidrcnn}              & 92.67/92.20          & 74.91/74.21          & 54.54/53.45      \\ \hline
PDV (Ours)              & \textbf{93.13/92.71} & \textbf{75.49/74.91} & 54.75/\textbf{53.90}      \\ \hline
\multirow{2}{*}{Method} & \multicolumn{3}{c}{LEVEL 2 mAP/mAPH}                           \\
                        & 0-30m                & 30-50m               & 50m-Inf          \\ \hline
$\textrm{PV-RCNN}^\star$ \cite{shi2020pv}                & 91.58/91.00          & 65.13/64.49          & 36.46/35.70      \\
Voxel-RCNN \cite{deng2020voxelrcnn}              & 91.74/-              & 67.89/-              & 40.80/-          \\
CT3D \cite{Sheng_2021_ICCV_ct3d}                    & 91.76/-              & 68.93/-              & \textbf{42.60}/-          \\ \hline
PDV (Ours)              & \textbf{92.41/91.99} & \textbf{69.36/68.81} & 42.16/\textbf{41.48}      \\ \hline
\end{tabular}
}
\caption{Performance comparison on the Waymo Open Dataset with 202 validation sequences for 3D vehicle detection across distance. $\star$: Results are on Waymo Open Dataset 1.0 version.}
\label{tab:waymo-dist}
\end{table}

\noindent\textbf{Datasets.} We evaluate \method on the Waymo Open Dataset~\cite{sun2020scalability} and the KITTI 3D Object Detection benchmark~\cite{geiger2012vision}. The Waymo Open Dataset is one of the biggest and most diverse autonomous driving datasets available, containing 798 training sequences (approximately 158k point cloud samples) and 202 validation sequences (approximately 40k point cloud samples), with annotations for objects in full 360$^{\circ}$ field of view. The Waymo Open Dataset uses standard mean average precision (mAP) as well as mAPH, which  factors in heading angle. The predictions are split into LEVEL\_1, which only includes 3D labels with more than five LiDAR points, and LEVEL\_2, which also includes 3D labels with at least one LiDAR point. The KITTI dataset contains 7,481 training samples and 7,518 testing samples and uses standard average precision (AP) on easy, moderate and hard difficulties. We adopt the standard \textit{training} and \textit{val} split sets from Chen et al.~\cite{chen20153d}.

\noindent\textbf{Input Parameters.} For the Waymo Open Dataset~\cite{sun2020scalability}, the detection range is [-75.2m, 75.2m] for the X and Y axes, and [-2m, 4m] for the Z axis. We divide the raw point cloud into voxels of size (0.1m, 0.1m, 0.15m). Since the KITTI dataset~\cite{geiger2012vision} only provides annotations in front camera's field of view, its detection range is set to be [0, 70.4m] for the X axis, [-40m, 40m] for the Y axis, and [-3m, 1m] for the Z axis. We set the voxel size to be (0.05m, 0.05m, 0.1m).

\noindent\textbf{Network Architecture.} \method uses the last two voxel layers $l=3,4$ for voxel point centroid localization. Density-aware RoI grid pooling uses a grid size of $U=6$. Each ball query uses a set of radii $r= \left[\left[0.8, 1.2\right], \left[1.2, 2.4\right]\right]$ to aggregate voxel point centroid features at each layer with two 32-layer FFNs for $l=3$ and two 64-layer FFNs for $l=4$. KDE for each ball query is calculated using a Gaussian kernel, $w$, with bandwidth $\sigma=0.25$. The self-attention module uses one transformer encoder layer with a single attention head. We follow Pointformer~\cite{pan20213d} for the FFN size for point density positional encoding with the added density feature from the number of points in each RoI grid voxel.

\noindent\textbf{Training and Inference Details.} \method is trained end-to-end with the Adam optimizer \cite{kingma2014adam}. We start with an initial learning rate of 0.01 and update it using one-cycle policy~\cite{smith2018disciplined} and cosine annealing~\cite{loshchilov2016sgdr}. We train the model for 50 epochs on the Waymo Open Dataset~\cite{sun2020scalability} with a batch size of 24 on 6 NVIDIA Tesla V100 GPUs and 80 epochs on the KITTI Dataset \cite{geiger2012vision} with a batch size of 4 on 2 NVIDIA Tesla P100 GPUs. We adopt commonly used data augmentation strategies for LiDAR 3D object detection, including random flipping about the X axis or Y axis, random global scaling with random scaling factor between 0.95 and 1.05, random global rotations about the Z axis between $-\frac{\pi}{4}$ and $\frac{\pi}{4}$, and ground truth data augmentation~\cite{second}. For post-processing, we use a non-maximum-suppresion (NMS) threshold of 0.1 for both Waymo and KITTI to remove redundant boxes.

\begin{table}[!t]
\centering
\begin{tabular}{l|ccc}
\hline
\multicolumn{1}{c|}{\multirow{2}{*}{Model}} & \multicolumn{3}{c}{LEVEL\_2 mAPH}                \\
\multicolumn{1}{c|}{}                       & Veh.           & Ped.           & Cyc.           \\ \hline
$\textrm{PV-RCNN}^\dagger$ \cite{shi2020pv}                                     & 68.41          & 57.61          & 63.98          \\
PV-RCNN++ \cite{shi2021pv}                                   & 69.71          & 59.72          & 65.17          \\ \hline
PDV (Ours)                                  & \textbf{69.98} & \textbf{60.00} & \textbf{67.88} \\ \hline
\end{tabular}
\caption{Performance comparison on the Waymo validation set for 3D multi-class with first and second LiDAR return. $\dagger$: Results are from~\cite{shi2021pv}.}
\label{tab:waymo-2nd-lidar}
\end{table}

\begin{table*}[t!]
\centering
\begin{tabular}{l|ccc|ccc|ccc}
\hline
                         & \multicolumn{3}{c|}{Car 3D (IoU = 0.7)}          & \multicolumn{3}{c|}{Pedestrian 3D (IoU = 0.5)}   & \multicolumn{3}{c}{Cyclist 3D (IoU = 0.5)}       \\
\multirow{-2}{*}{Method} & Easy           & Moderate       & Hard           & Easy           & Moderate       & Hard           & Easy           & Moderate       & Hard           \\ \hline
$\textrm{PV-RCNN}^\star$ \cite{shi2020pv}                  & 92.10          & 84.36          & 82.48          & 64.26          & 56.67          & 51.91          & 88.88          & 71.95          & 66.78          \\
$\textrm{CT3D}^\star$ \cite{Sheng_2021_ICCV_ct3d}                     & 92.34          & 84.97          & 82.91          & 61.05          & 55.57          & 51.10          & 89.01          & 71.88          & 67.91          \\ \hline
PDV (Ours)               & \textbf{92.56} & \textbf{85.29} & \textbf{83.05} & \textbf{66.90} & \textbf{60.80} & \textbf{55.85} & \textbf{92.72} & \textbf{74.23} & \textbf{69.60} \\
\rowcolor[HTML]{E0FFFF} 
\textit{Improvement}     & \textit{+0.22} & \textit{+0.32} & \textit{+0.14} & \textit{+2.64} & \textit{+4.13} & \textit{+3.94} & \textit{+3.71} & \textit{+2.28} & \textit{+1.69} \\ \hline
\end{tabular}
\caption{3D detection results on the KITTI \textit{val} set for car, pedestrian, and cyclist classes using $\textrm{AP}|_{R_{40}}$. $\star$: Results are taken from publicly released models \cite{shi2020pv, Sheng_2021_ICCV_ct3d, openpcdet2020}.}
\label{tab:kitti-val}
\end{table*}

\begin{table}[ht!]
\centering
\footnotesize
\resizebox{\columnwidth}{!}{
\begin{tabular}{l|ccc|ccc}
\hline
\multirow{2}{*}{Method} & \multicolumn{3}{c|}{Car 3D (IoU=0.7)}            & \multicolumn{3}{c}{Cyclist 3D (IoU=0.5)}         \\
                        & Easy           & Mod.           & Hard           & Easy           & Mod.           & Hard           \\ \hline
SECOND\cite{second}                  & 83.34          & 72.55          & 65.82          & 71.33          & 52.08          & 45.83          \\
PointPillar\cite{lang2019pointpillars}             & 82.58          & 74.31          & 68.99          & 77.10          & 58.65          & 51.92          \\
STD\cite{yang2019std}                     & 87.95          & 79.71          & 75.09          & 78.69          & 61.59          & 55.30          \\
3DSSD\cite{Yang20203dssd}                   & 88.36          & 79.57          & 74.55          & {\ul 82.48}    & 64.10          & 56.90          \\
SA-SSD\cite{he2020sassd}                  & 88.75          & 79.79          & 74.16          & -              & -              & -              \\
SE-SSD\cite{sessd}                  & \textbf{91.49} & \textbf{82.54} & 77.15          & -              & -              & -              \\
PV-RCNN\cite{shi2020pv}                 & 90.25          & 81.43          & 76.82          & 78.60          & 63.71          & 57.65          \\
PV-RCNN++\cite{shi2021pv}               & 90.14          & 81.88          & 77.15          & 82.22          & 67.33          & 60.04          \\
Voxel-RCNN\cite{deng2020voxelrcnn}              & {\ul 90.90}    & 81.62          & 77.06          & -              & -              & -              \\
DSA-PV\cite{bhattacharyya2021sa}                  & 88.25          & 81.46          & 76.96          & 82.19          & \textbf{68.54} & \textbf{61.33} \\
CT3D\cite{Sheng_2021_ICCV_ct3d}                    & 87.83          & 81.77          & 77.16          & -              & -              & -              \\
VoTr-TSD\cite{Mao_2021_ICCV_votr}                & 89.90          & {\ul 82.09}    & \textbf{79.14} & -              & -              & -              \\
Pyramid-PV\cite{Mao_2021_ICCV_pyramidrcnn}              & 88.39          & 82.08          & {\ul 77.49}    & -              & -              & -              \\ \hline
PDV (Ours)              & 90.43          & 81.86          & 77.36          & \textbf{83.04} & {\ul 67.81}    & {\ul 60.46}    \\ \hline
\end{tabular}
}
\caption{3D detection results on the KITTI \textit{test} set for car and cyclist using $\textrm{AP}|_{R_{40}}$. Bolded and underlined values are best and second-best performance, respectively.}
\label{tab:kitti-test}
\end{table}

\subsection{Waymo Dataset Results}
We report the multi-class Waymo Open Dataset results on the validation set in~\Cref{tab:waymo-val}. \method achieves state-of-the-art results on all classes on both LEVEL\_1 and LEVEL\_2 mAP/mAPH metrics. We outperform methods that use PV-RCNN as a base architecture~\cite{shi2020pv, Mao_2021_ICCV_votr, Mao_2021_ICCV_pyramidrcnn} by at least +2.13\% on vehicle LEVEL\_2 mAPH. \method performs well on the other classes as well, increasing performance by +2.99\% and +1.88\%  on pedestrian and cyclist LEVEL\_2 3D mAPH, respectively, compared to PV-RCNN. We also outperform PV-RCNN++~\cite{shi2021pv} by +1.25\%, +0.46\%, and +0.71\% on vehicle, pedestrian, and cyclist LEVEL\_2 3D mAPH, respectively. \method effectively captures fine point details lost in the voxel backbone through point density for accurate bounding box refinement in the second stage.

\Cref{tab:waymo-dist} shows \method's performance across distance on the Waymo Open Dataset for the vehicle class. \method outperforms all methods by +0.46\%/+0.51\% and +0.42\%/+0.70\% at 0-30m and 30-50m for LEVEL\_1 mAP/mAPH, respectively, and +0.65\%/+0.99\% and +0.43\%/+4.32\% for LEVEL\_2 mAP/mAPH, respectively, better utilizing Waymo's close-range LiDAR data for accurate detection. We also report the results on the Waymo validation set using first and second LiDAR return in \Cref{tab:waymo-2nd-lidar}, surpassing both PV-RCNN and PV-RCNN++. Detailed second LiDAR return results can be found in the supplementary materials.

\subsection{KITTI Dataset Results}
\Cref{tab:kitti-val} shows the results on the KITTI \textit{val} set. \method achieves state-of-the-art multi-class results, improving 3D $\textrm{AP}|_{R_{40}}$ performance by +0.32\%, +4.13\%, and +2.28\% on car, pedestrian, and cyclist classes, respectively, on the moderate difficulty. As shown in \Cref{tab:kitti-test}, \method also obtains competitive results on the KITTI \textit{test} set, showing improvements over PV-RCNN~\cite{shi2020pv}, Voxel-RCNN~\cite{deng2020voxelrcnn}, and CT3D~\cite{Sheng_2021_ICCV_ct3d} with an increase of at least +0.09\% on the moderate 3D $\textrm{AP}|_{R_{40}}$ car class, but falling short compared to other methods~\cite{sessd, Mao_2021_ICCV_votr, Mao_2021_ICCV_pyramidrcnn, shi2021pv}. We hypothesize that since detection architectures can use a higher voxel resolution for the truncated KITTI point cloud, \method's modules provide smaller improvements compared to other methods.

\subsection{Ablation Studies}
We provide ablation studies to study the effects of each component in \method. All models and variations are trained on 10\% of the Waymo training dataset for 60 epochs.

\noindent\textbf{Components.} \Cref{tab:waymo-ab-components} shows the relative performance gain of each component on LEVEL\_2 mAPH on the Waymo Open Dataset validation set. Experiment~\ref{itm:comp1} uses voxel centers to localize voxel features. By localizing voxel features using voxel point centroids instead, Experiment~\ref{itm:comp2} provides an improvement of +0.83\%, +3.86\%, and +3.67\% for the vehicle, pedestrian, and cyclist classes, respectively. Voxel feature localization is extremely beneficial for smaller objects since voxel centers do not have proper alignment with the point cloud. By localizing the features closer to the object's scanned surface, voxel point centroids contain meaningful geometric shape information for proposal refinement.

Local feature density estimation improves scores by +0.07\%, 4.45\%, +1.78\% as shown in Experiment~\ref{itm:comp3}. By capturing feature density relationships via KDE, local feature density estimation provides a large improvement for the deformable pedestrian class, where encoded features have variable spatial configurations. Experiment~\ref{itm:comp4} shows attention improves results by +0.45\%, +0.61\%, +1.43\% providing a large increase in the pedestrian and cyclist class by establishing long-range dependencies between RoI grid points. Finally, Experiment~\ref{itm:comp5} shows the effect of density confidence prediction, improving performance on the pedestrian and cyclist classes by +0.23\% and +0.74\%, respectively, with a small drop on the vehicle class of -0.01\%.

\noindent\textbf{Point Density Positional Encoding.} \Cref{tab:waymo-ab-pe} shows the effects of point density positional encoding in density-aware RoI pooling. Experiment~\ref{itm:pe1} shows the baseline performance without any positional encoding. Experiment~\ref{itm:pe2} uses a sinusoidal encoding similar to DETR~\cite{carion2020end, vaswani2017attention} where each spatial dimension is encoded with independent sine and cosine functions at different frequencies, showing an improvement on the pedestrian and cyclist by +0.14\% and +0.50\%, but a reduction on vehicle by -0.25\%. A FFN with spatial grid point coordinates as input provides an overall increase in performance of +0.17\%, +0.18\%, and +0.27\% on vehicles, pedestrians, and cyclists, respectively, as shown in Experiment~\ref{itm:pe3}. In Experiment~\ref{itm:pe4}, the density feature, in the form of number of points within each grid point voxel, shows similar improvements over the baseline with +0.22\%, +0.73\%, and +0.67\%. When we combine both local spatial coordinates and density in Experiment~\ref{itm:pe5}, we obtain the best performance with increases of +0.15\%, +1.07\%, +1.82\%.

\begin{table}[!t]
\centering
\resizebox{\columnwidth}{!}{
\begin{tabular}{c|cccl|ccc}
\hline
\multirow{2}{*}{Exp.} & \multirow{2}{*}{VC}  & \multirow{2}{*}{LD} & \multirow{2}{*}{GA}   & \multicolumn{1}{c|}{\multirow{2}{*}{DC}} & \multicolumn{3}{c}{LEVEL\_2 mAPH}                \\
                      &                      &                      &                      & \multicolumn{1}{c|}{}                    & Veh.           & Ped.           & Cyc.           \\ \hline
\newtag{1}{itm:comp1}                     & \multicolumn{1}{l}{} & \multicolumn{1}{l}{} & \multicolumn{1}{l}{} &                                          & 64.01          & 43.15          & 56.41          \\
\newtag{2}{itm:comp2}                     & \checkmark                    & \multicolumn{1}{l}{} & \multicolumn{1}{l}{} &                                          & 64.84          & 47.01          & 60.08          \\
\newtag{3}{itm:comp3}                     & \checkmark                    & \checkmark                    & \multicolumn{1}{l}{} &                                          & 64.91          & 51.46          & 61.86          \\
\newtag{4}{itm:comp4}                     & \checkmark                    & \checkmark                    & \checkmark                    &                                          & \textbf{65.36} & 52.07          & 63.29          \\
\newtag{5}{itm:comp5}                     & \checkmark                    & \checkmark                    & \checkmark                    & \multicolumn{1}{c|}{\checkmark}                   & 65.35          & \textbf{52.30} & \textbf{64.03} \\ \hline
\end{tabular}
}
\caption{\method ablation experiments trained on 10\% of the Waymo training set. VC indicates voxel point centroids, LD indicates local feature density estimation, GA indicates grid point self-attention, and DC indicates density confidence prediction.}
\label{tab:waymo-ab-components}
\end{table}

\begin{table}[!ht]
\centering
\begin{tabular}{c|c|ccc}
\hline
\multirow{2}{*}{Exp.} & \multirow{2}{*}{PE} & \multicolumn{3}{c}{LEVEL\_2 mAPH}                \\
                      &                     & Veh.           & Ped.           & Cyc.           \\ \hline
\newtag{1}{itm:pe1}                     & None                & 65.20          & 51.23          & 62.21          \\
\newtag{2}{itm:pe2}                     & Sinusoidal          & 64.95          & 51.37          & 62.71          \\
\newtag{3}{itm:pe3}                     & FFN (XYZ)           & 65.37          & 51.41          & 62.48          \\
\newtag{4}{itm:pe4}                     & FFN (D)             & \textbf{65.42} & 51.96          & 62.88          \\
\newtag{5}{itm:pe5}                     & FFN (XYZD)          & 65.35          & \textbf{52.30} & \textbf{64.03} \\ \hline
\end{tabular}
\caption{Positional encoding (PE) ablation experiments trained on 10\% of the Waymo training dataset. XYZ indicates spatial grid locations and D indicates number of points in each grid voxel.}
\label{tab:waymo-ab-pe}
\end{table}

\subsection{Runtime Analysis}
\Cref{tab:runtime} shows the runtime comparisons of \method. We train the models for 60 epochs on 10\% of the training data for Waymo and 80 epochs on the \textit{training} set for KITTI. Each model is evaluated with an Intel i7-6850K processor, a single Titan Xp GPU, and a batch size of 1. For a fair comparison, we use the same number of $K$ proposals from the RPN ($K=275$ for Waymo and $K=100$ for KITTI) for all models. \method outperforms PV-RCNN on inference speed and performance, reducing runtime by 14\% and 5\% for Waymo and KITTI, respectively, and improving mAPH by +2.42\% and mAP by +1.51\% on Waymo and KITTI, respectively.

\Cref{tab:runtime} also shows the issue with voxel-based methods on larger input spaces. Although Voxel-RCNN is computationally efficient and performs well on the KITTI dataset, its performance degrades significantly, performing worse than PV-RCNN by -0.67\% on Waymo. On the other hand, voxel point centroids provide an efficient alternative that retains spatial fidelity across the two datasets. Using voxel point centroids only (VC Only) results in an efficient architecture with a decrease in runtime by 61\% and better performance on LEVEL\_2 mAPH by +1.22\% compared to PV-RCNN on Waymo. VC Only also scales to the large Waymo dataset, with only an increase of 64 ms in runtime from KITTI, showing voxel point centroid localization as an efficient alternative to FPS for spatially locating voxel features. \method overcomes the limitations of voxel-based methods when scaling to larger datasets, while retaining computational efficiency.
\begin{table}[!t]
\centering
\small
\resizebox{\columnwidth}{!}{
\begin{tabular}{l|cc|cc}
\hline
\multirow{2}{*}{Method} & \multicolumn{2}{c|}{Waymo}   & \multicolumn{2}{c}{KITTI}    \\
                        & Speed (ms)  & mAPH           & Speed (ms)  & 3D mAP         \\ \hline
$\textrm{PV-RCNN}^\star$ \cite{shi2020pv}                & 396         & 58.14          & 142         & 71.93          \\
$\textrm{Voxel-RCNN}^\star$ \cite{deng2020voxelrcnn}             & \textbf{91} & 57.47          & \textbf{74} & 72.97          \\ \hline
VC Only (Ours)          & 154         & 59.36          & 90          & 72.82          \\
PDV (Ours)              & 340         & \textbf{60.56} & 135         & \textbf{73.44} \\ \hline
\end{tabular}}
\caption{Runtime comparison and performance on the Waymo and KITTI validation sets. We average the LEVEL\_2 3D mAPH for Waymo and moderate difficulty 3D AP for KITTI across classes. $\star$: Models are trained using the publicly released code\cite{openpcdet2020}.}
\label{tab:runtime}
\end{table}

\begin{figure}[!t]
\centering
    \includegraphics[width=0.9\columnwidth]{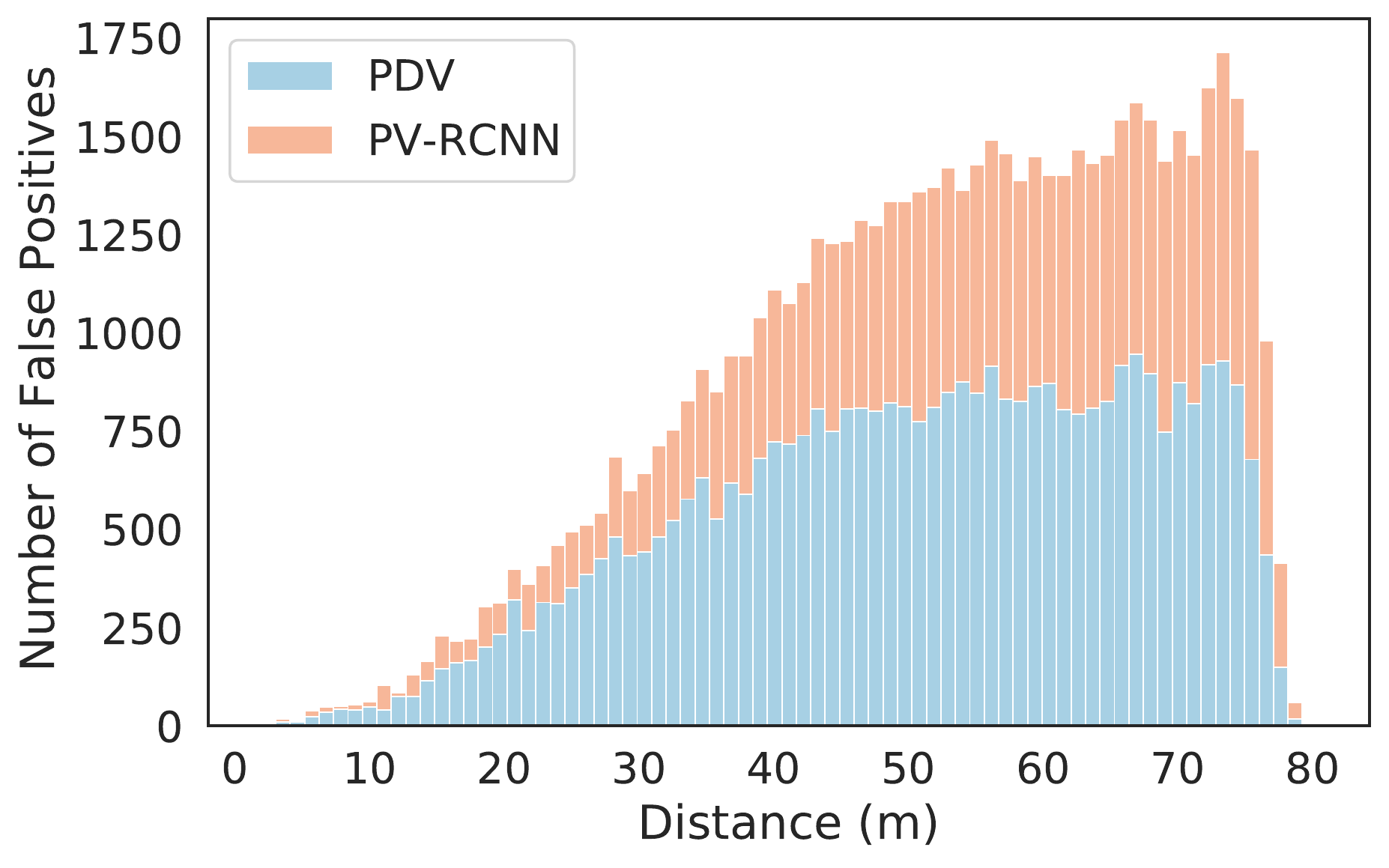}
\caption{Number of false positive predictions on the vehicle class ($\textrm{IoU}<0.7$) across distance for \method and PV-RCNN evaluated on 10\% of the Waymo validation set.}
\label{fig:waymo-fps}
\end{figure}

\subsection{False Positives across Distance}
\Cref{fig:waymo-fps} shows the number of false positives ($\textrm{IoU}<0.7$) predicted by \method and PV-RCNN across distance for vehicles on the Waymo dataset. As the distance from the sensor increases, the gap between \method and PV-RCNN increases. We attribute the increased differential to using point density to refine bounding box regression and confidence values, which is more beneficial at detecting challenging objects at farther ranges.
\section{Conclusion}
\label{sec:conclusion}

We present \method, a novel LiDAR 3D object detection method that uses voxel features and raw point cloud data to account for point density variations in LiDAR point clouds. \method is particularly useful on large input spaces where point cloud sampling is expensive and voxel resolutions are low, resulting in state-of-the-art performance on the Waymo dataset and competitive results for the KITTI dataset.
\clearpage
{\small
\bibliographystyle{ieee_fullname}
\bibliography{pdv}
}
\clearpage
\appendix

In the supplementary material, we provide a discussion on the potential negative impact and limitations of \method in \Cref{sec:sm-neg-impact} and \Cref{sec:sm-limitations}, respectively. We also provide additional detailed results on the Waymo Open Dataset~\cite{sun2020scalability} in \Cref{sec:sm-waymo} as well as on the KITTI dataset~\cite{geiger2012vision} in \Cref{sec:sm-kitti}. Finally, we provide visualizations of the point density variations across distance in \Cref{sec:sm-pdd}.

\section{Potential Negative Impact}
\label{sec:sm-neg-impact}
As autonomous driving progresses towards public use, \method has the potential to be used for surveillance of civilians. Privacy is a large concern in the digital age, and it is important to ensure that this information is not used without consent or to maliciously track people's location.

Unlike other domains where safety concerns are minimal, reliable 3D object detection in autonomous driving is extremely important. Objects that are missed or even misclassified could have detrimental effects in subsequent decision-making tasks for the vehicle, running a risk for both the passengers and other individuals sharing the road. It is important to take the results of 3D object detection methods on improved benchmarks as only one aspect for reliable 3D object detection as increased performance on a metric is not indicative of the practicality of a method in a real-world scenario.

\section{Limitations}
\label{sec:sm-limitations}
\noindent\textbf{Voxel Resolution.} \method shows the most improvement compared to other methods on large input spaces when the voxel resolution is limited, such as on the Waymo Open Dataset~\cite{sun2020scalability}. However, when the voxel resolution is sufficiently high, the improvements from \method are less significant, as is the case on the KITTI dataset~\cite{geiger2012vision}. At higher voxel resolutions, the difference between the voxel center and the voxel point centroids decreases, resulting in less performance gains. Additionally, at high voxel resolutions, each occupied voxel represents a decent approximation of the point density, which can be captured through standard convolutions. Thus, \method is most suited for when low voxel resolutions are necessary, such as for 360\textdegree~detection, where the point density feature can be fully exploited for better performance.

\noindent\textbf{LiDAR Dependency.} \method uses point density as an additional encoding feature, and therefore relies on the LiDAR specifications for accurate 3D object detection. Although \method is tested on the KITTI dataset~\cite{geiger2012vision} and the Waymo Open Dataset~\cite{sun2020scalability}, which have different LiDAR characteristics, it is necessary to expand other datasets to ensure that \method is robust to different LiDAR sampling patterns. Some examples include Nuscenes~\cite{caesar2020nuscenes}, which uses a 32 beam LiDAR, and Cirrus~\cite{wang2021cirrus}, which uses a LiDAR designed for long range detection.

\noindent\textbf{Adverse Weather Conditions.} \method uses point density to provide additional encoding in the second stage. If there is a significant distribution shift in the LiDAR point density during inference time, there may be significant degradation in performance. For example, as highlighted in SPG~\cite{xu2021spg}, the Waymo Kirkland dataset~\cite{sun2020scalability} has a different distribution of points, where rainy weather results in a significant number of points missing on objects. Since \method relies on point density, a sudden shift in point density variations due to weather may cause severe degradation in performance. A future improvement to \method is to ensure robustness to potential shifts in point density distributions during inference.

\section{More Waymo Dataset Results}
\label{sec:sm-waymo}

We show the Waymo Open Dataset validation results for the pedestrian and cyclist classes across distance for \method using the first LiDAR return only in \Cref{tab:sm-waymo-dist-ped} and \Cref{tab:sm-waymo-dist-cyc}, respectively. The distance evaluation is broken down into three categories: 0 to 30 meters, 30 to 50 meters, and beyond 50 meters. We also show the multi-class results for \method using first and second LiDAR return in \Cref{tab:sm-waymo-val-2nd}. Overall, \method utilizes the additional points from the second LiDAR return better than PV-RCNN++, providing the largest performance increase on the cyclist class.

\begin{table}[!t]
\centering
\resizebox{\columnwidth}{!}{
\begin{tabular}{l|ccc}
\hline
\multirow{2}{*}{Method} & \multicolumn{3}{c}{Pedestrian LEVEL 1 mAP/mAPH} \\
                        & 0-30m        & 30-50m       & 50m-Inf     \\ \hline
PDV (Ours)              & 80.32/73.60  & 72.97/63.28  & 61.69/50.07 \\ \hline
\multirow{2}{*}{Method} & \multicolumn{3}{c}{Pedestrian LEVEL 2 mAP/mAPH} \\
                        & 0-30m        & 30-50m       & 50m-Inf     \\ \hline
PDV (Ours)              & 75.26/68.82  & 65.78/56.85  & 47.46/38.30 \\ \hline
\end{tabular}
}
\caption{Performance comparison on the Waymo Open Dataset with 202 validation sequences for 3D pedestrian (IoU~=~0.5) detection across distance.}
\label{tab:sm-waymo-dist-ped}
\end{table}

\begin{table}[!t]
\centering
\resizebox{\columnwidth}{!}{
\begin{tabular}{l|ccc}
\hline
\multirow{2}{*}{Method} & \multicolumn{3}{c}{Cyclist LEVEL 1 mAP/mAPH} \\
                        & 0-30m        & 30-50m       & 50m-Inf     \\ \hline
PDV (Ours)              & 80.86/79.83  & 62.61/61.45  & 46.23/44.12 \\ \hline
\multirow{2}{*}{Method} & \multicolumn{3}{c}{Cyclist LEVEL 2 mAP/mAPH} \\
                        & 0-30m        & 30-50m       & 50m-Inf     \\ \hline
PDV (Ours)              & 80.42/79.40  & 58.95/57.87  & 43.05/41.09 \\ \hline
\end{tabular}
}
\caption{Performance comparison on the Waymo Open Dataset with 202 validation sequences for 3D cyclist (IoU~=~0.5) detection across distance.}
\label{tab:sm-waymo-dist-cyc}
\end{table}

\begin{table*}[!t]
\centering
\small
\resizebox{\textwidth}{!}{
\begin{tabular}{l|cccc|cccc|cccc}
\hline
                         & \multicolumn{2}{c}{Veh. (LEVEL\_1)} & \multicolumn{2}{c|}{Veh. (LEVEL\_2)} & \multicolumn{2}{c}{Ped. (LEVEL\_1)} & \multicolumn{2}{c|}{Ped. (LEVEL\_2)} & \multicolumn{2}{c}{Cyc. (LEVEL\_1)} & \multicolumn{2}{c}{Cyc. (LEVEL\_2)} \\
\multirow{-2}{*}{Method} & mAP              & mAPH             & mAP               & mAPH             & mAP              & mAPH             & mAP               & mAPH             & mAP              & mAPH             & mAP              & mAPH             \\ \hline
$\textrm{PV-RCNN}^\dagger$ \cite{shi2020pv}                  & 77.51            & 76.89            & 68.98             & 68.41            & 75.01            & 65.65            & 66.04             & 57.61            & 67.81            & 66.35            & 65.39            & 63.98            \\
PV-RCNN++ \cite{shi2021pv}                & 78.79            & 78.21            & 70.26             & 69.71            & 76.67            & 67.15            & \textbf{68.51}             & 59.72            & 68.98            & 67.63            & 66.48            & 65.17            \\ \hline
PDV (Ours)               & \textbf{79.43}   & \textbf{78.89}   & \textbf{70.47}    & \textbf{69.98}   & \textbf{76.94}   & \textbf{68.09}   & 68.07    & \textbf{60.00}   & \textbf{71.45}   & \textbf{70.18}   & \textbf{69.11}   & \textbf{67.88}   \\
\rowcolor[HTML]{E0FFFF} 
\textit{Improvement}     & \textit{+0.64}   & \textit{+0.68}   & \textit{+0.21}    & \textit{+0.27}   & \textit{+0.27}   & \textit{+0.94}   & \textit{-0.44}    & \textit{+0.28}   & \textit{+2.47}   & \textit{+2.55}   & \textit{+2.63}   & \textit{+2.71}   \\ \hline
\end{tabular}
}
\caption{Performance comparison on the Waymo Open Dataset with 202 validation sequences for 3D vehicle (IoU~=~0.7), pedestrian (IoU~=~0.5) and cyclist (IoU~=~0.5) detection using first and second LiDAR return. $\dagger$: Results are from~\cite{shi2021pv}.}
\label{tab:sm-waymo-val-2nd}
\end{table*}

\section{More KITTI Dataset Results}
\label{sec:sm-kitti}

We provide KITTI dataset results on the \textit{val} set on 3D and BEV for  $\textrm{AP}|_{R_{11}}$ and $\textrm{AP}|_{R_{40}}$ in \Cref{tab:sm-kitti-3d-r11}, \Cref{tab:sm-kitti-bev-r11}, and \Cref{tab:sm-kitti-bev-r40}. \method's performance is compared to PV-RCNN~\cite{shi2020pv} and CT3D~\cite{Sheng_2021_ICCV_ct3d}.

\begin{table*}[!t]
\centering
\begin{tabular}{l|ccc|ccc|ccc}
\hline
                         & \multicolumn{3}{c|}{Car 3D ($R_{11}$)}          & \multicolumn{3}{c|}{Pedestrian 3D ($R_{11}$)}   & \multicolumn{3}{c}{Cyclist 3D ($R_{11}$)}       \\
\multirow{-2}{*}{Method} & Easy           & Moderate       & Hard           & Easy           & Moderate       & Hard           & Easy           & Moderate       & Hard           \\ \hline
$\textrm{PV-RCNN}^\star$ \cite{shi2020pv}                  & 89.35          & 83.69          & 78.70          & 64.60           & 57.90           & 53.23          & 85.22          & 70.47          & 65.75          \\
$\textrm{CT3D}^\star$ \cite{Sheng_2021_ICCV_ct3d}                     & 89.11          & \textbf{85.04} & 78.76          & 61.74          & 56.28          & 52.51          & 85.04          & 71.71          & 68.05          \\ \hline
PDV (Ours)               & \textbf{89.52} & 84.03          & \textbf{79.09} & \textbf{65.83} & \textbf{61.18} & \textbf{55.87} & \textbf{90.48} & \textbf{73.23} & \textbf{69.55} \\
\rowcolor[HTML]{E0FFFF} 
\textit{Improvement}     & \textit{+0.41} & \textit{-1.01} & \textit{+0.33} & \textit{+1.23} & \textit{+3.28} & \textit{+2.64} & \textit{+5.26} & \textit{+1.52} & \textit{+1.50}  \\ \hline
\end{tabular}
\caption{3D detection results on the KITTI \textit{val} set for car, pedestrian, and cyclist classes using $\textrm{AP}|_{R_{11}}$. $\star$: Results are taken from publicly released models \cite{shi2020pv, Sheng_2021_ICCV_ct3d, openpcdet2020}.}
\label{tab:sm-kitti-3d-r11}
\end{table*}

\begin{table*}[!t]
\centering
\begin{tabular}{l|ccc|ccc|ccc}
\hline
                         & \multicolumn{3}{c|}{Car BEV ($R_{11}$)}         & \multicolumn{3}{c|}{Pedestrian BEV ($R_{11}$)}  & \multicolumn{3}{c}{Cyclist BEV ($R_{11}$)}      \\
\multirow{-2}{*}{Method} & Easy           & Moderate       & Hard           & Easy           & Moderate       & Hard           & Easy           & Moderate       & Hard           \\ \hline
$\textrm{PV-RCNN}^\star$ \cite{shi2020pv}                  & 90.09          & 87.90          & 87.41          & 67.01          & 61.38          & 56.10           & 86.79          & 73.55          & 69.69          \\
$\textrm{CT3D}^\star$ \cite{Sheng_2021_ICCV_ct3d}                     & 90.25          & 88.18          & 87.78          & 64.23          & 59.84          & 55.76          & \textbf{90.94} & 73.68          & \textbf{71.21} \\ \hline
PDV (Ours)               & \textbf{90.33} & \textbf{88.33} & \textbf{87.91} & \textbf{69.01} & \textbf{63.54} & \textbf{59.46} & 90.77          & \textbf{73.75} & \textbf{71.21} \\
\rowcolor[HTML]{E0FFFF} 
\textit{Improvement}     & \textit{+0.08} & \textit{+0.15} & \textit{+0.13} & \textit{+2.00}    & \textit{+2.16} & \textit{+3.36} & \textit{-0.17} & \textit{+0.07} & \textit{+0.00}    \\ \hline
\end{tabular}
\caption{BEV detection results on the KITTI \textit{val} set for car, pedestrian, and cyclist classes using $\textrm{AP}|_{R_{11}}$. $\star$: Results are taken from publicly released models \cite{shi2020pv, Sheng_2021_ICCV_ct3d, openpcdet2020}.}
\label{tab:sm-kitti-bev-r11}
\end{table*}

\begin{table*}[!t]
\centering
\begin{tabular}{l|ccc|ccc|ccc}
\hline
                         & \multicolumn{3}{c|}{Car BEV ($R_{40}$)}         & \multicolumn{3}{c|}{Pedestrian BEV ($R_{40}$)}  & \multicolumn{3}{c}{Cyclist BEV ($R_{40}$)}      \\
\multirow{-2}{*}{Method} & Easy           & Moderate       & Hard           & Easy           & Moderate       & Hard           & Easy           & Moderate       & Hard           \\ \hline
$\textrm{PV-RCNN}^\star$ \cite{shi2020pv}                  & 93.02          & 90.33          & 88.53          & 67.97          & 60.52          & 55.80           & 91.02          & 74.54          & 69.92          \\
$\textrm{CT3D}^\star$ \cite{Sheng_2021_ICCV_ct3d}                     & \textbf{95.92} & \textbf{91.35} & 89.29          & 64.41          & 59.18          & 54.86          & 92.60          & 75.40          & 71.31          \\ \hline
PDV (Ours)               & 93.60          & 91.14          & \textbf{90.74} & \textbf{69.40} & \textbf{63.42} & \textbf{58.70} & \textbf{93.09} & \textbf{76.08} & \textbf{71.46} \\
\rowcolor[HTML]{E0FFFF} 
\textit{Improvement}     & \textit{-2.32} & \textit{-0.21} & \textit{+1.45} & \textit{+1.43} & \textit{+2.90}  & \textit{+2.90}  & \textit{+0.49} & \textit{+0.68} & \textit{+0.15} \\ \hline
\end{tabular}
\caption{BEV detection results on the KITTI \textit{val} set for car, pedestrian, and cyclist classes using $\textrm{AP}|_{R_{40}}$. $\star$: Results are taken from publicly released models \cite{shi2020pv, Sheng_2021_ICCV_ct3d, openpcdet2020}.}
\label{tab:sm-kitti-bev-r40}
\end{table*}

\section{Point Density Distance Plots}
\label{sec:sm-pdd}
\begin{figure*}[!t]
\centering

\begin{tabular}{ccc}
    \includegraphics[width=0.64\columnwidth]{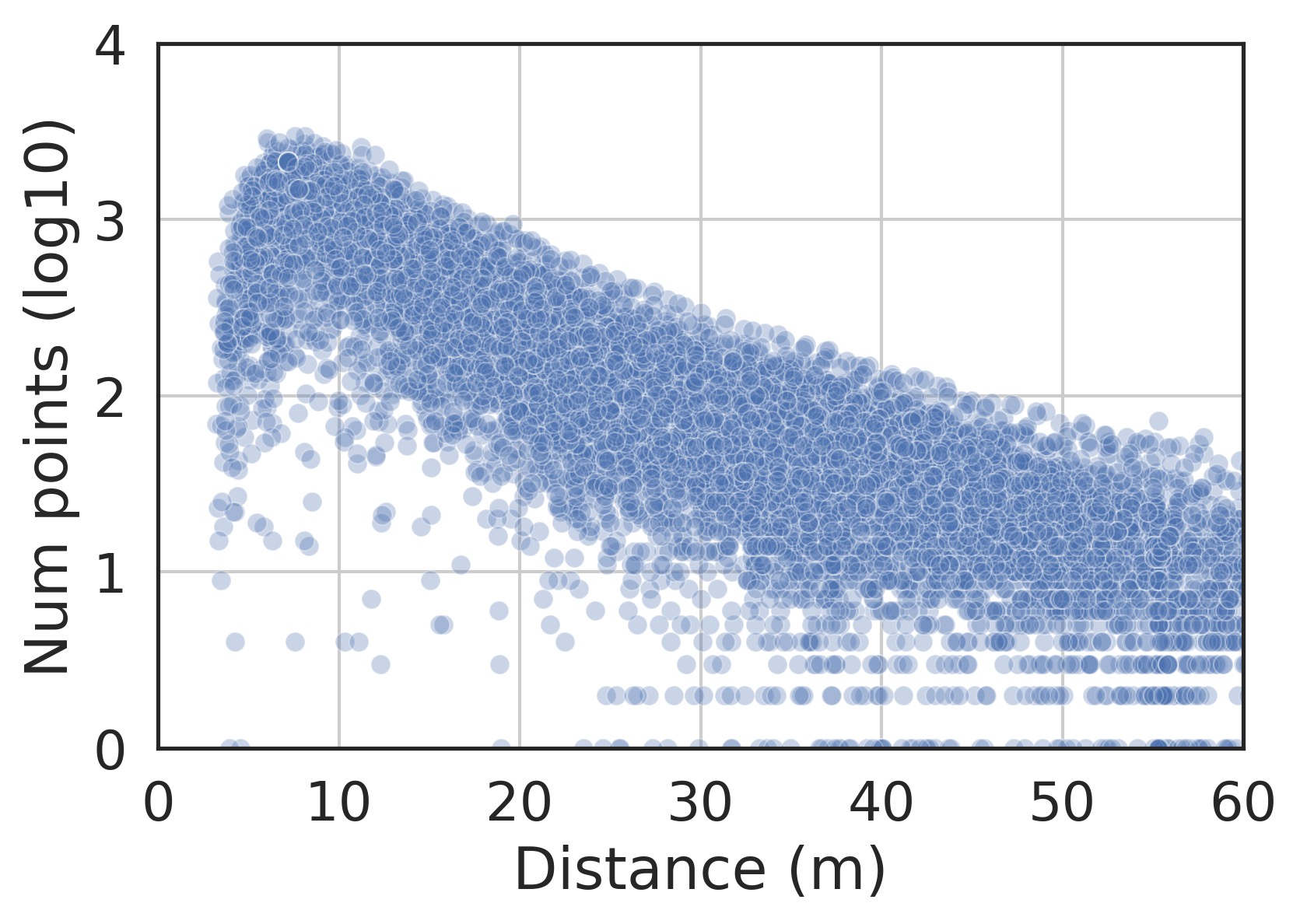} &
    \includegraphics[width=0.64\columnwidth]{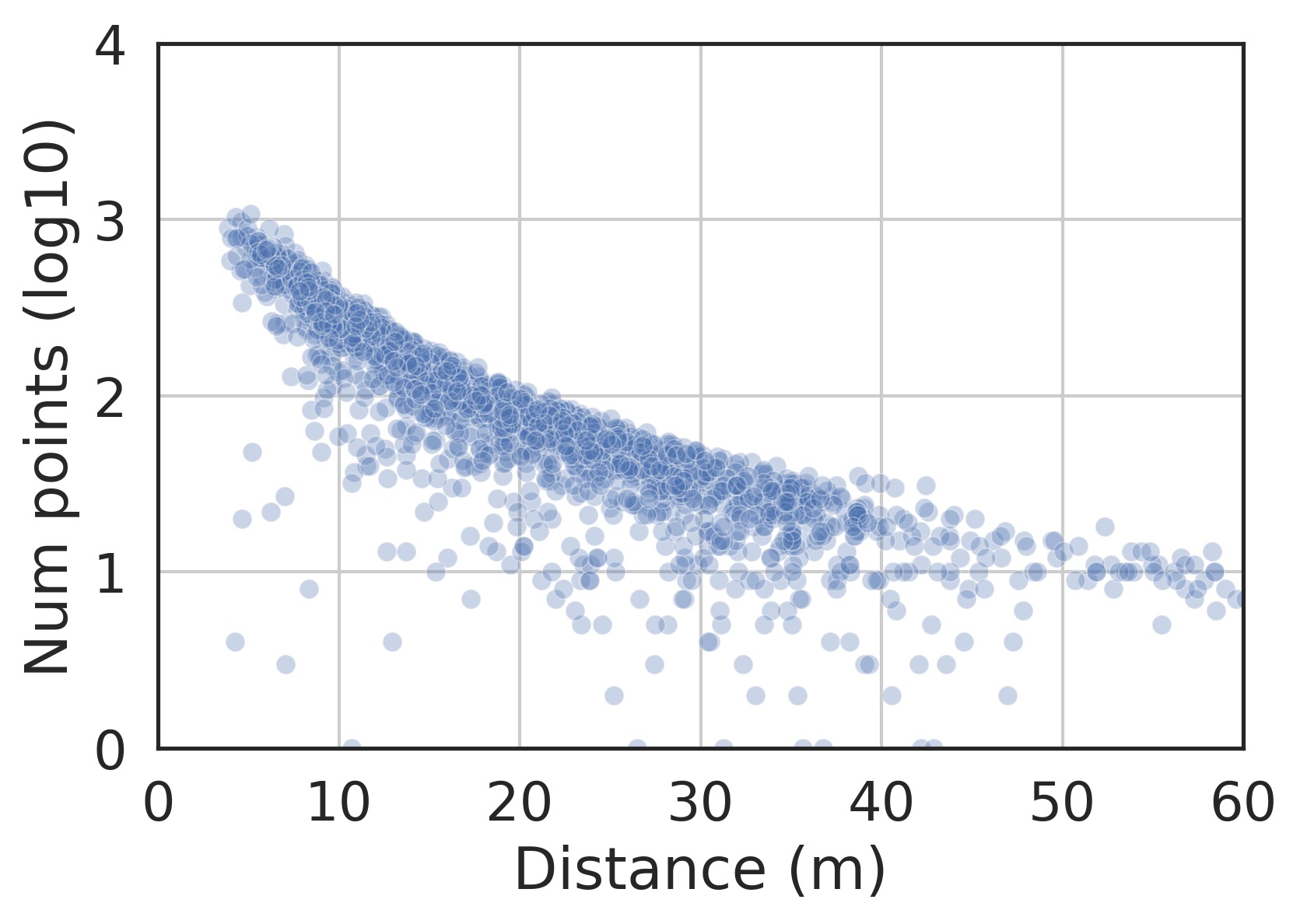} &
    \includegraphics[width=0.64\columnwidth]{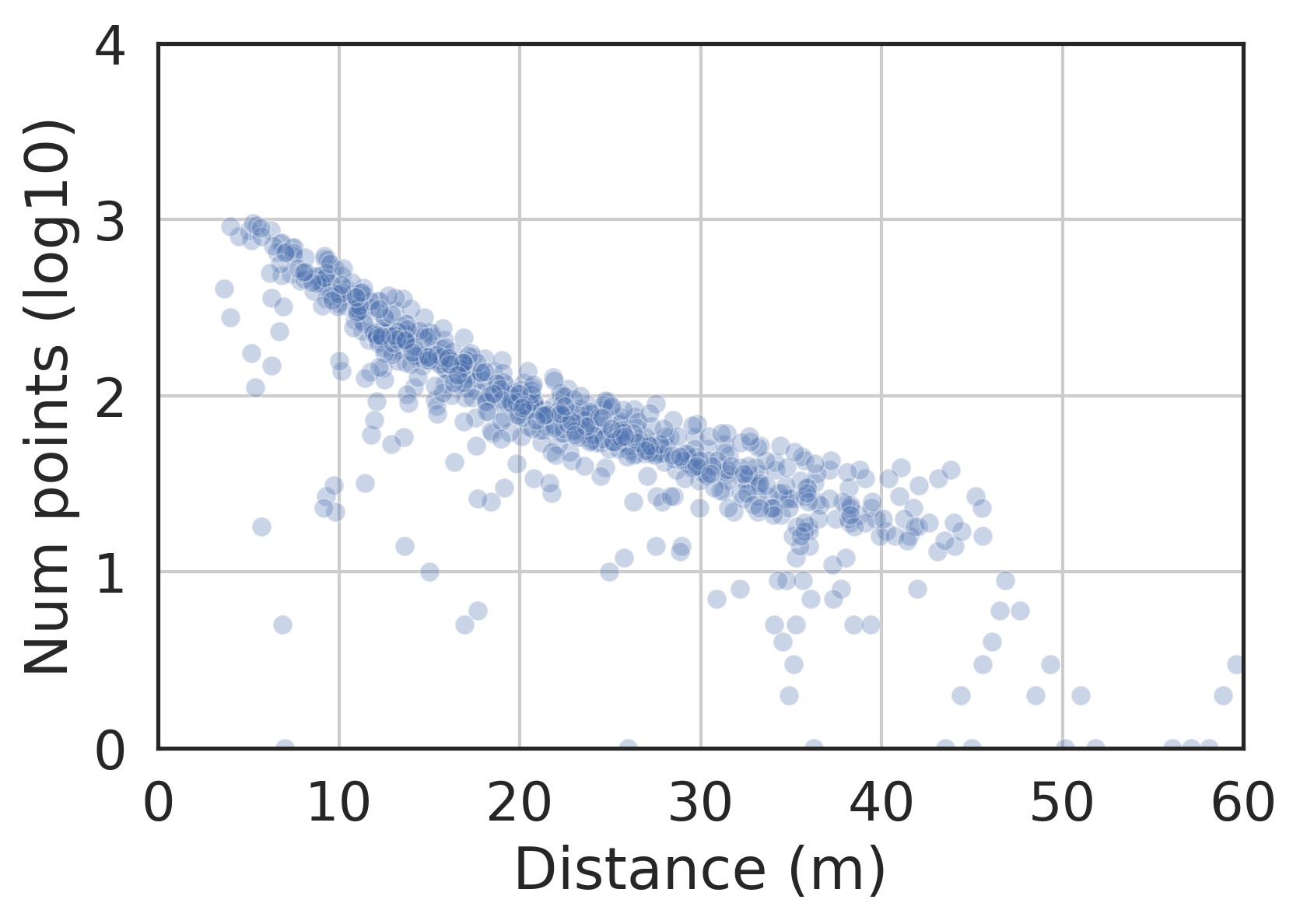}\\
    (a) & (b) & (c)\\
    \includegraphics[width=0.64\columnwidth]{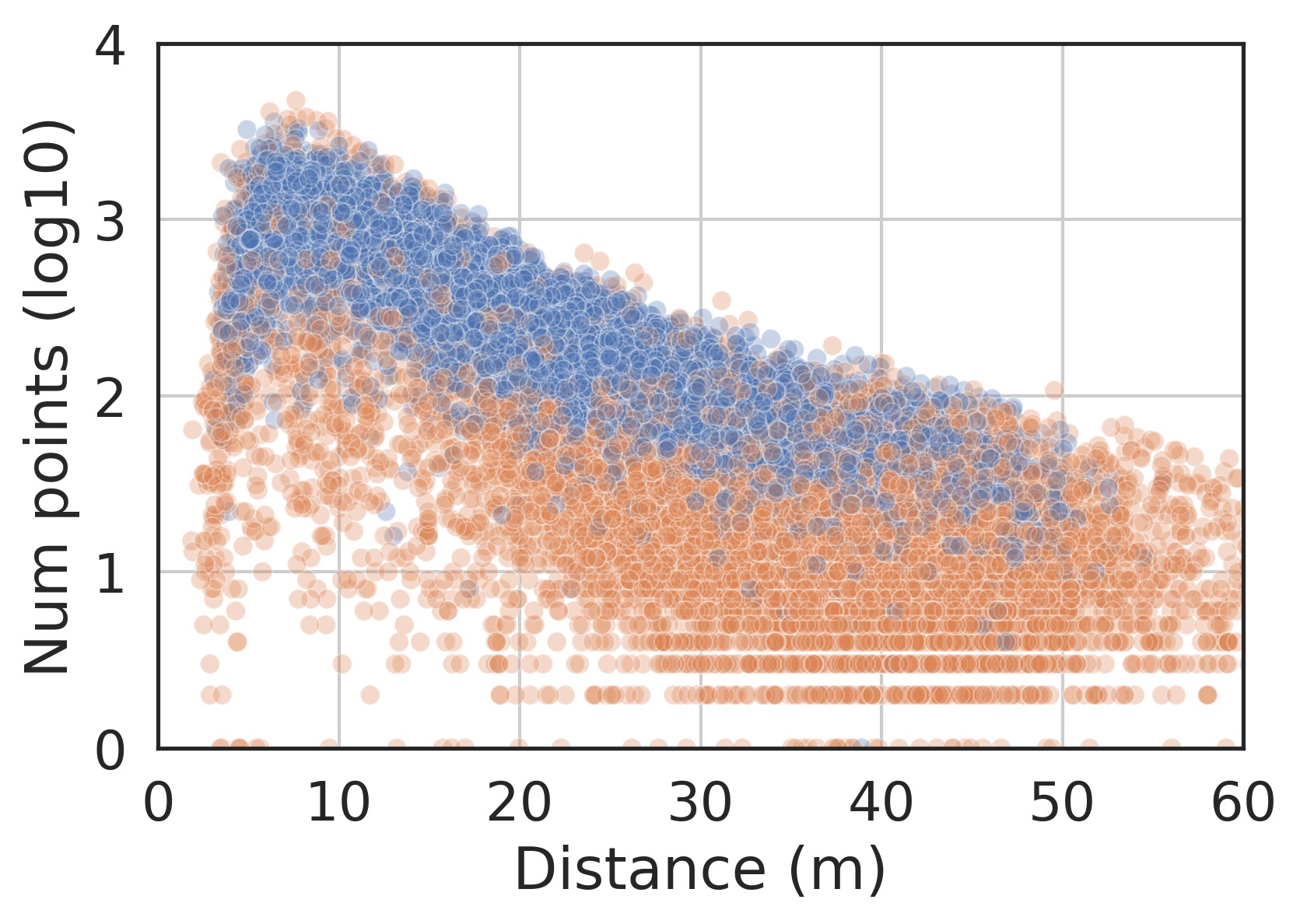} &
    \includegraphics[width=0.64\columnwidth]{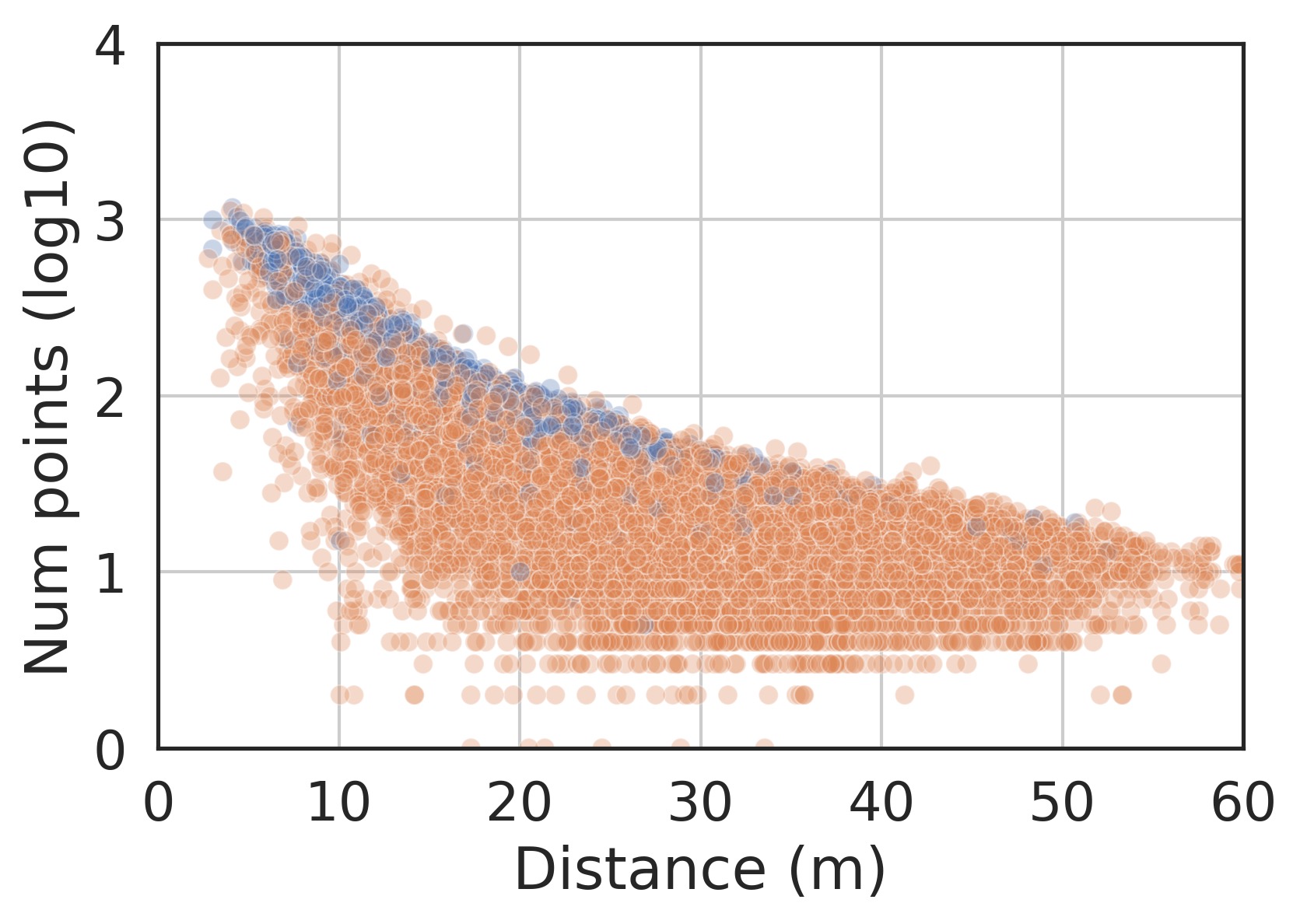} &
    \includegraphics[width=0.64\columnwidth]{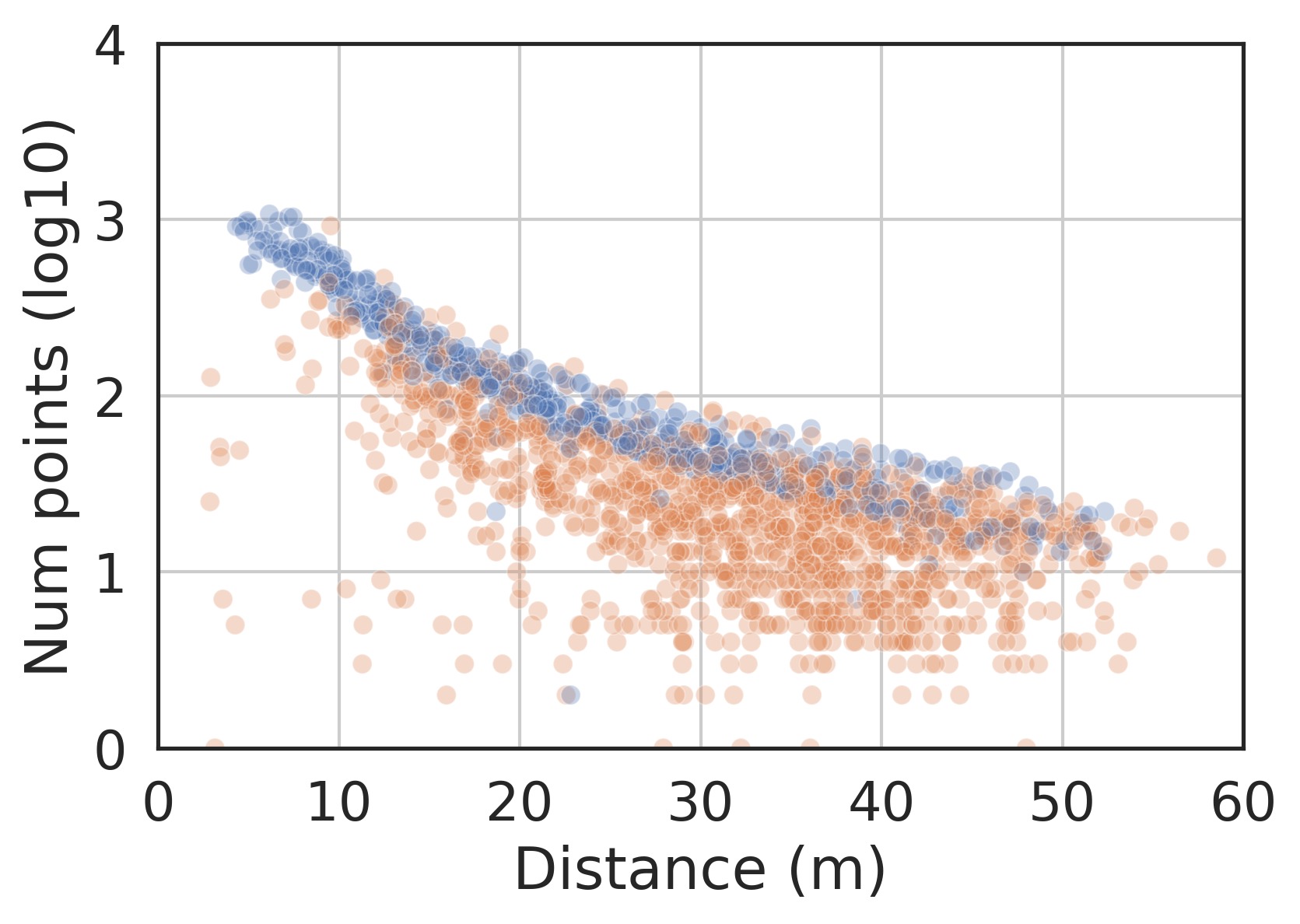}\\
    (d) & (e) & (f)\\
    \includegraphics[width=0.64\columnwidth]{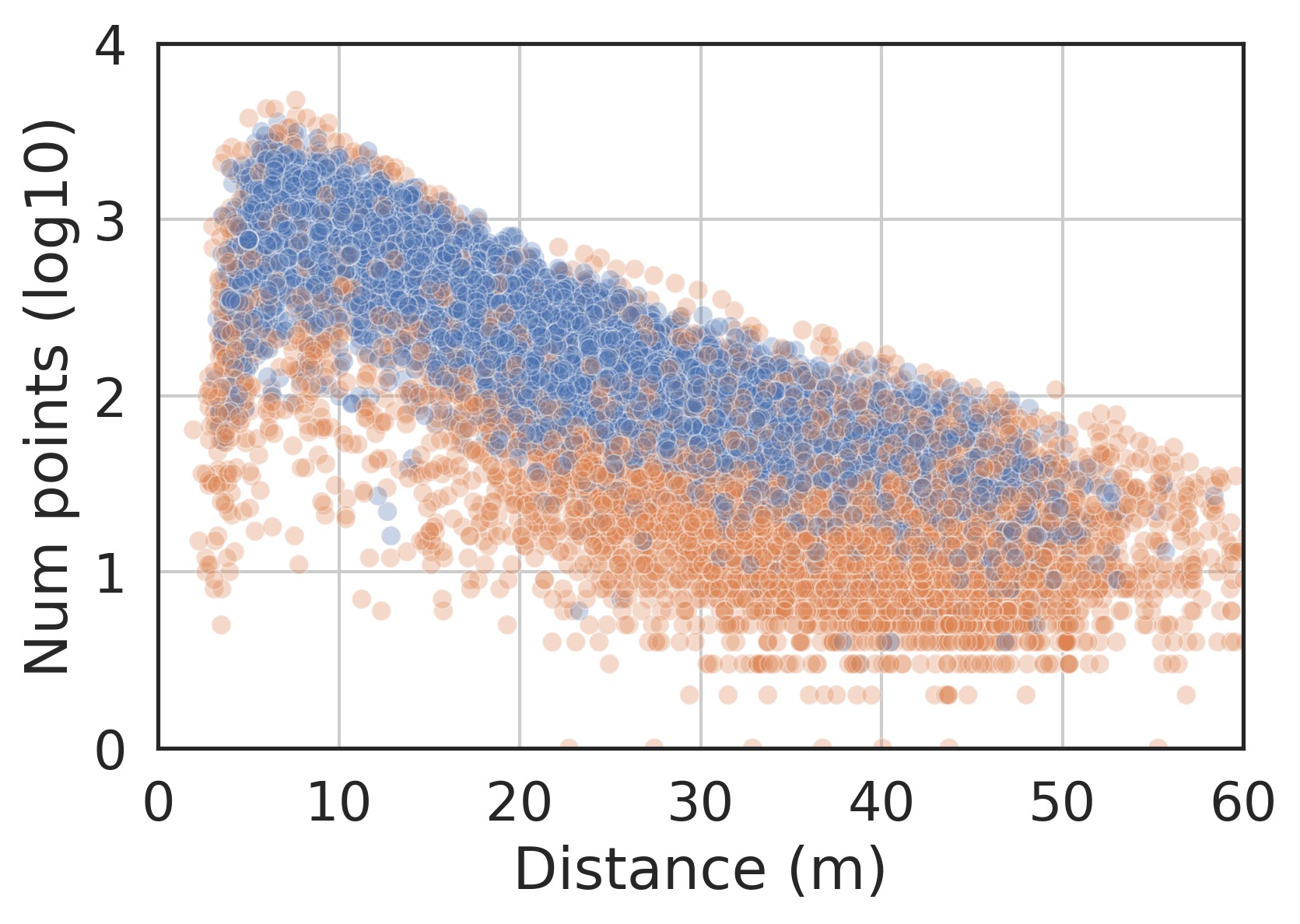} &
    \includegraphics[width=0.64\columnwidth]{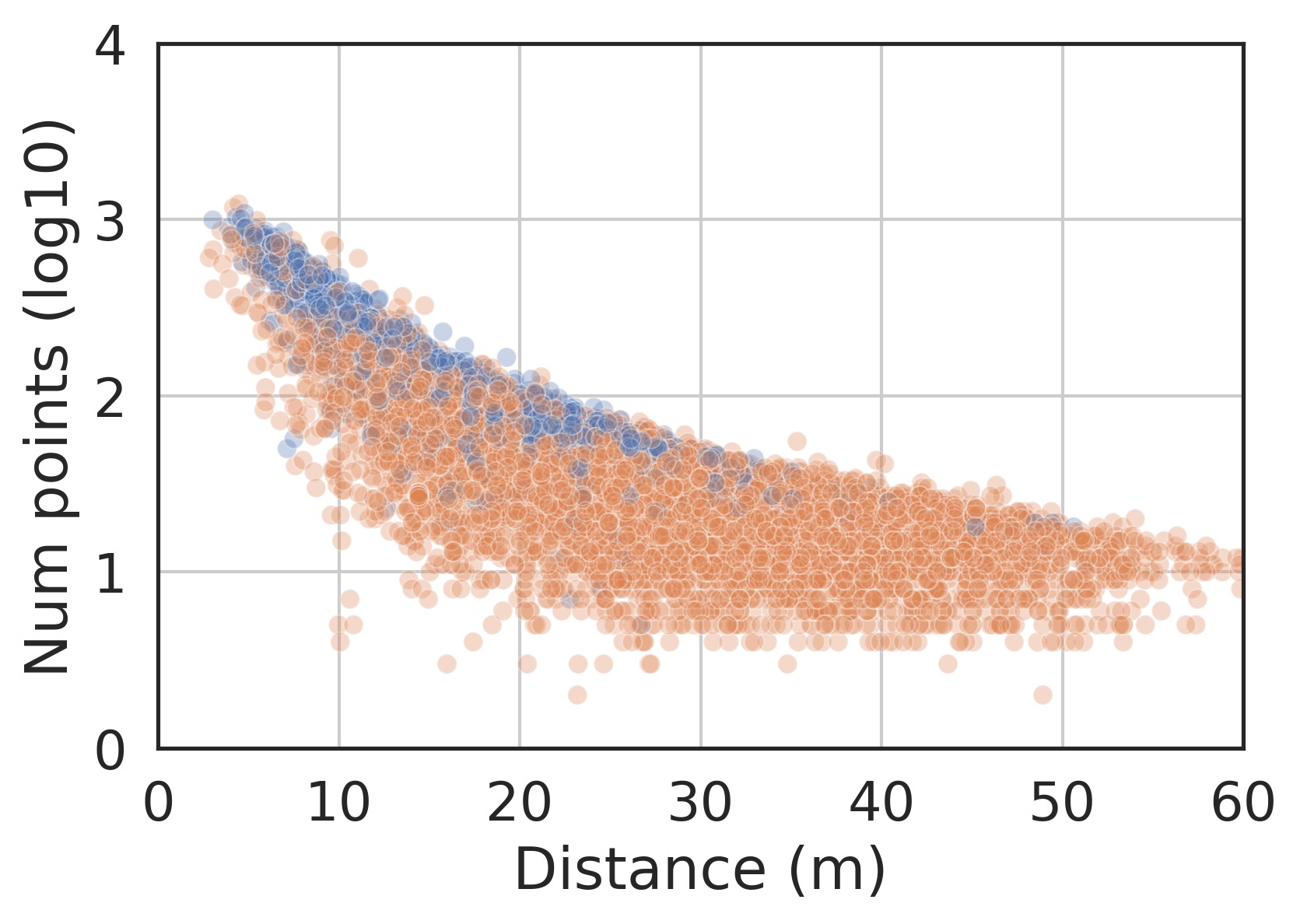} &
    \includegraphics[width=0.64\columnwidth]{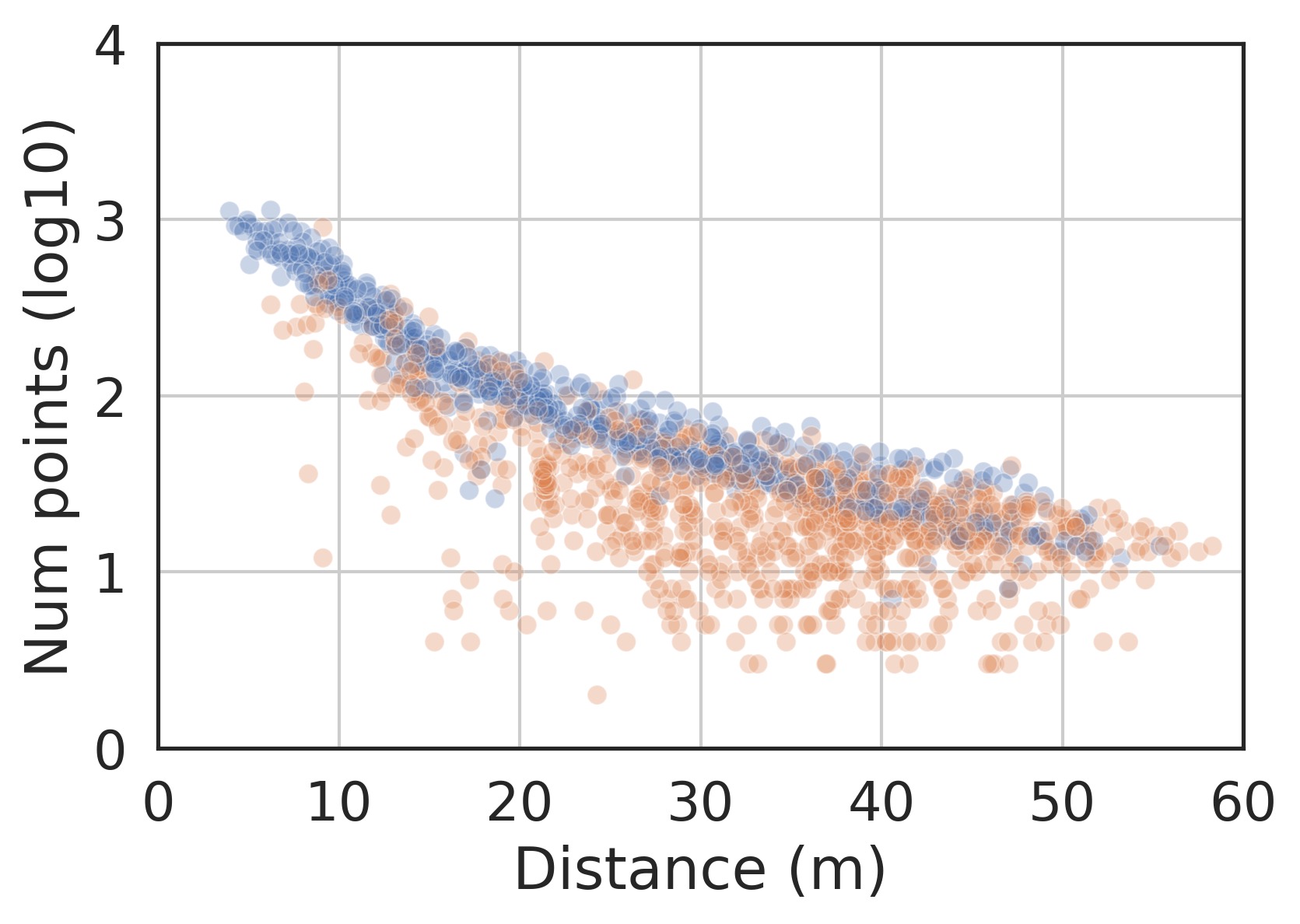}\\
    (g) & (h) & (i)\\
\end{tabular}
\caption{From left to right column: Number of points in ground truth boxes across distance for cars, pedestrians and cyclists on the KITTI dataset. The first row shows the distribution of training samples on the \textit{train} split. The second and third row show the predictions of PV-RCNN~\cite{shi2020pv} and \method on the KITTI \textit{val} split, respectively. \textcolor{blue}{Blue} predictions are true positives while \textcolor{orange}{orange} predictions are false positives for cars ($\textrm{IoU}<0.7$), pedestrians ($\textrm{IoU}<0.5$), and cyclists ($\textrm{IoU}<0.5$).}
\label{fig:sm-kitti-pdd}
\end{figure*}

\Cref{fig:sm-kitti-pdd} shows point density across distance plots for PV-RCNN~\cite{shi2020pv} and \method. By using the relationship between distance and point density (number of points within each final bounding box prediction), \method effectively reduces the number of false positives outside the distribution of training samples across distance.

\end{document}